\documentclass[10pt]{article}
\usepackage{float}
\usepackage{xcolor}
\usepackage{lineno}
\usepackage{adjustbox}
\usepackage{amssymb}
\usepackage[inline]{enumitem}
\let\savedcontentsline\addcontentsline
\usepackage[preprint]{style}
\setcitestyle{numbers,square,comma}
\let\addcontentsline\savedcontentsline

\usepackage{amsmath,amsfonts,bm}

\def\secref#1{section~\ref{#1}}

\def\eqref#1{equation~\ref{#1}}

\def\1{\bm{1}}

\DeclareMathAlphabet{\mathsfit}{\encodingdefault}{\sfdefault}{m}{sl}
\SetMathAlphabet{\mathsfit}{bold}{\encodingdefault}{\sfdefault}{bx}{n}

\usepackage{amsmath}
\usepackage{graphicx}
\usepackage{wrapfig}
\usepackage{float}
\usepackage{algorithm}
\usepackage{algpseudocode}
\usepackage{tikz}
\usepackage{booktabs}
\usepackage[most]{tcolorbox}
\usetikzlibrary{arrows.meta,calc,decorations.pathreplacing}

\newtheorem{theorem}{Theorem}

\tcbset{statement/.style={
  enhanced,
  breakable,
  colback=gray!10,
  frame hidden,
  boxrule=0pt,
  sharp corners,
  left=5pt,
  right=5pt,
  top=4pt,
  bottom=4pt,
  before skip=6pt,
  after skip=6pt
}}
\tcolorboxenvironment{proposition}{statement}
\tcolorboxenvironment{theorem}{statement}

\usepackage[utf8]{inputenc} 
\usepackage[T1]{fontenc}    
\usepackage{hyperref}       
\usepackage{url}            
\newcommand{\doi}[1]{doi: \href{https://doi.org/#1}{\nolinkurl{#1}}}
\usepackage{amsfonts}       
\usepackage{nicefrac}       
\usepackage{microtype}      
\usepackage{titletoc}
\titlecontents{section}[1.8em]
  {\addvspace{6pt}\normalsize\bfseries}
  {\contentslabel{1.8em}}{}
  {\hfill\contentspage}
\titlecontents{subsection}[4.3em]
  {\small}
  {\contentslabel{2.5em}}{}
  {\titlerule*[0.6em]{.}\contentspage}
\hypersetup{
  colorlinks=true,
  linkcolor=blue!65!black,
  citecolor=blue!65!black,
  urlcolor=blue!55!black
}

\makeatletter
\def\input@path{{../}{../figures/}{figures/}}
\makeatother
\graphicspath{{../figures/}{figures/}{../}{../../}}

\title{Function-Space Certification for Neural Operators}

\author{
    Ryan Hsiang$^{1,2}$ \quad
    Robert Joseph George$^{2}$ \quad
    Anima Anandkumar$^{2}$\thanks{Corresponding author.} \\[4pt]
    $^{1}$Department of Electrical Engineering, National Taiwan University \\
    $^{2}$Department of Computing and Mathematical Sciences, California Institute of Technology
}

\begin{document}

\maketitle

\begin{abstract}
Certifying the robustness and physical consistency of scientific machine-learning models ensures that their predictions remain reliable beyond the test set. For neural operators, which learn PDE solution operators by mapping infinite-dimensional function spaces, certification means establishing rigorous, domain-wide guarantees for their output functions. However, standard neural network (NN) verification methods are fundamentally designed for vector-to-vector mappings, and they can only certify neural operators at fixed grid resolutions and provide no guarantees over the entire domain or across another resolution. We close this gap by introducing a function-space certification framework that verifies neural operators across the entire continuous domain. We formulate interval bound propagation (IBP) and CROWN in function space, propagating upper and lower bound functions through each layer using rigorous numerical integration. In addition, we extend bound propagation to higher-order derivatives to certify PDE residuals and solution error bounds. Experiments across several neural operator architectures and scientific problems, including the 572.5-million-parameter spherical Fourier neural operator for global weather forecasting, show that our method produces tight, domain-wide guarantees efficiently, providing a foundation for trustworthy operator learning.
\end{abstract}

\section{Introduction}
\begin{figure}[t]
    \centering
    \resizebox{\linewidth}{!}{%
    \begin{tikzpicture}[
        >=Latex,
        font=\footnotesize,
        title/.style={font=\small\bfseries, align=center},
        layer/.style={
            draw=gray!70!black,
            fill=gray!8,
            rounded corners=2pt,
            thick,
            align=center,
            minimum height=0.8cm
        },
        spectral/.style={
            draw=magenta!70!red!75!black,
            fill=magenta!70!red!16,
            rounded corners=2pt,
            thick,
            align=center,
            minimum height=0.9cm
        },
        integ/.style={
            draw=magenta!70!red!75!black,
            fill=magenta!70!red!9,
            rounded corners,
            thick,
            align=center,
            text width=5.3cm,
            inner sep=4pt
        },
        prop/.style={
            draw=blue!65!black,
            fill=blue!8,
            rounded corners,
            thick,
            align=center,
            minimum height=1.0cm
        },
        app/.style={
            draw=gray!60,
            fill=gray!5,
            rounded corners=2pt,
            align=center,
            minimum height=0.85cm,
            inner xsep=4pt
        },
        flow/.style={->, thick},
        bigflow/.style={->, line width=1.3pt, gray!70!black},
        declare function={
            ain(\x)=0.95+0.45*sin(2*pi*\x/2.8 r)+0.15*cos(4*pi*\x/2.8 r);
            uout(\x)=0.95-0.35*cos(2*pi*\x/2.8 r)+0.12*sin(4*pi*\x/2.8 r);
            uw(\x)=0.13+0.06*sin(6*pi*\x/2.8 r)^2;
        }
    ]

    \node[title] at (1.4,2.6) {Input function set $\mathcal A$};
    \fill[orange!20]
        plot[domain=0:2.8,samples=80] (\x,{ain(\x)+0.17})
        -- plot[domain=2.8:0,samples=80] (\x,{ain(\x)-0.17})
        -- cycle;
    \draw[orange!80!black,thick]
        plot[domain=0:2.8,samples=80] (\x,{ain(\x)+0.17});
    \draw[orange!80!black,thick]
        plot[domain=0:2.8,samples=80] (\x,{ain(\x)-0.17});
    \draw[thick]
        plot[domain=0:2.8,samples=120]
        (\x,{ain(\x)+0.12*sin(9*pi*\x/2.8 r)});
    \draw[->] (0,0) -- (3.0,0) node[right] {$x$};
    \draw[->] (0,0) -- (0,2.1);
    \node[orange!80!black,anchor=west] at (1.55,1.9) {$\overline a(x)$};
    \node[orange!80!black,anchor=west] at (0.1,0.3) {$\underline a(x)$};
    \node[align=center] at (1.45,-0.55)
        {$\underline a(x)\le a(x)\le\overline a(x)$};

    \begin{scope}[shift={(4.0,0)}]
        \node[title] at (3.95,2.6)
            {Function-space IBP / CROWN};

        \node[layer, minimum width=0.7cm] (P) at (0.15,1.15) {$\mathcal P$};
        \node[layer, minimum width=2.9cm] (W) at (2.75,1.7)
            {$W_\ell z_\ell(x)+b_\ell$};
        \node[spectral, minimum width=2.9cm] (K) at (2.75,0.6)
            {$\displaystyle\int_\Omega \kappa_\ell(x-y)\,z_\ell(y)\,dy$};
        \node[circle,draw,thick,inner sep=1pt] (plus) at (4.75,1.15) {$+$};
        \node[layer, minimum width=0.7cm] (S) at (5.65,1.15) {$\sigma_\ell$};
        \node[layer, minimum width=0.7cm] (Q) at (7.55,1.15) {$\mathcal Q$};

        \draw[densely dashed,gray!70!black,rounded corners]
            (0.95,0.0) rectangle (6.2,2.3);
        \node[gray!40!black,anchor=south east,font=\scriptsize]
            at (6.2,0.0) {$\times\,n$ layers};

        \draw[bigflow] ($(P.west)+(-0.85,0)$) -- (P.west);
        \draw[flow] (P.east) -- ++(0.3,0) |- (W.west);
        \draw[flow] (P.east) -- ++(0.3,0) |- (K.west);
        \draw[flow] (W.east) -| (plus.north);
        \draw[flow] (K.east) -| (plus.south);
        \draw[flow] (plus) -- (S);
        \draw[flow] (S) -- (Q);
        \draw[bigflow] (Q.east) -- ++(0.85,0);

        \node[integ] (int) at (3.95,-0.85) {
            \textbf{Certified integration:} each integral is enclosed
            with a rigorous error bound
        };
        \draw[flow,magenta!70!red!75!black] (int.north -| K.south) -- (K.south);
    \end{scope}

    \begin{scope}[shift={(13.0,0)}]
        \node[title] at (1.4,2.6) {Certified output bounds};
        \fill[blue!15]
            plot[domain=0:2.8,samples=80] (\x,{uout(\x)+uw(\x)})
            -- plot[domain=2.8:0,samples=80] (\x,{uout(\x)-uw(\x)})
            -- cycle;
        \draw[blue!65!black,thick]
            plot[domain=0:2.8,samples=80] (\x,{uout(\x)+uw(\x)});
        \draw[blue!65!black,thick]
            plot[domain=0:2.8,samples=80] (\x,{uout(\x)-uw(\x)});
        \draw[thick]
            plot[domain=0:2.8,samples=80]
            (\x,{uout(\x)+0.05*sin(5*pi*\x/2.8 r)});
        \draw[->] (0,0) -- (3.0,0) node[right] {$x$};
        \draw[->] (0,0) -- (0,2.1);
        \node[blue!65!black,anchor=west] at (1.95,1.8) {$\overline u(x)$};
        \node[blue!65!black,anchor=west] at (0.1,0.22) {$\underline u(x)$};
        \node[align=center] at (1.45,-0.55)
            {$\underline u(x)\le \mathcal G_\theta(a)(x)\le\overline u(x)$};
    \end{scope}

    \node[title,anchor=west] at (0,-2.0) {Certified properties};
    \node[prop, minimum width=3.4cm, anchor=west] (rob) at (0,-2.95)
        {\textbf{Adversarial robustness}\\ output bounds over $\mathcal A$};
    \node[prop, minimum width=3.2cm, anchor=west] (res) at (3.9,-2.95)
        {\textbf{PDE residual}\\ $\|\mathcal L_a[u_\theta]\|\le\varepsilon_r$};
    \node[prop, minimum width=4.4cm, anchor=west] (bc) at (7.6,-2.95)
        {\textbf{Boundary / initial conditions}\\
         $\|\mathcal B_a[u_\theta]\|\le\delta_b,\quad
          \|\mathcal I_a[u_\theta]\|\le\delta_0$};
    \node[prop, minimum width=3.0cm, fill=blue!18, anchor=west] (sol)
        at (13.3,-2.95)
        {\textbf{Solution error}\\ $\|u_\theta-u\|\le\delta_u$};
    \draw[flow] (bc.east) -- (sol.west)
        node[midway,above,font=\scriptsize] {stability};
    \draw[flow] (res.north) -- ++(0,0.3) -| ([xshift=-0.4cm]sol.north);

    \node[title,anchor=west] at (0,-3.95) {Applications};
    \node[app,anchor=west] (a1) at (0,-4.75) {2D Poisson};
    \node[app,anchor=west] (a2) at ([xshift=0.2cm]a1.east) {1D Helmholtz};
    \node[app,anchor=west] (a3) at ([xshift=0.2cm]a2.east)
        {1D Burgers\\(space--time)};
    \node[app,anchor=west] (a4) at ([xshift=0.2cm]a3.east)
        {Kuramoto--Sivashinsky\\(long rollouts)};
    \node[app,anchor=west] (a5) at ([xshift=0.2cm]a4.east)
        {Global weather forecasting, 572.5M-parameter SFNO\\
         (maximum regional temperature)};
    \end{tikzpicture}%
    }
    \caption{
    \textbf{Overview of function-space certification for neural operators.}
    Function-space IBP and CROWN propagate the input through the neural operator. Each spectral convolution is an integral, which is
    enclosed with a rigorous error bound. The resulting upper and lower bound
    functions, and their derivatives, hold at every point of the domain and
    certify robustness, PDE residuals, boundary/initial conditions, and the solution error through a stability estimate.
    }
    \label{fig:overview}
\end{figure}
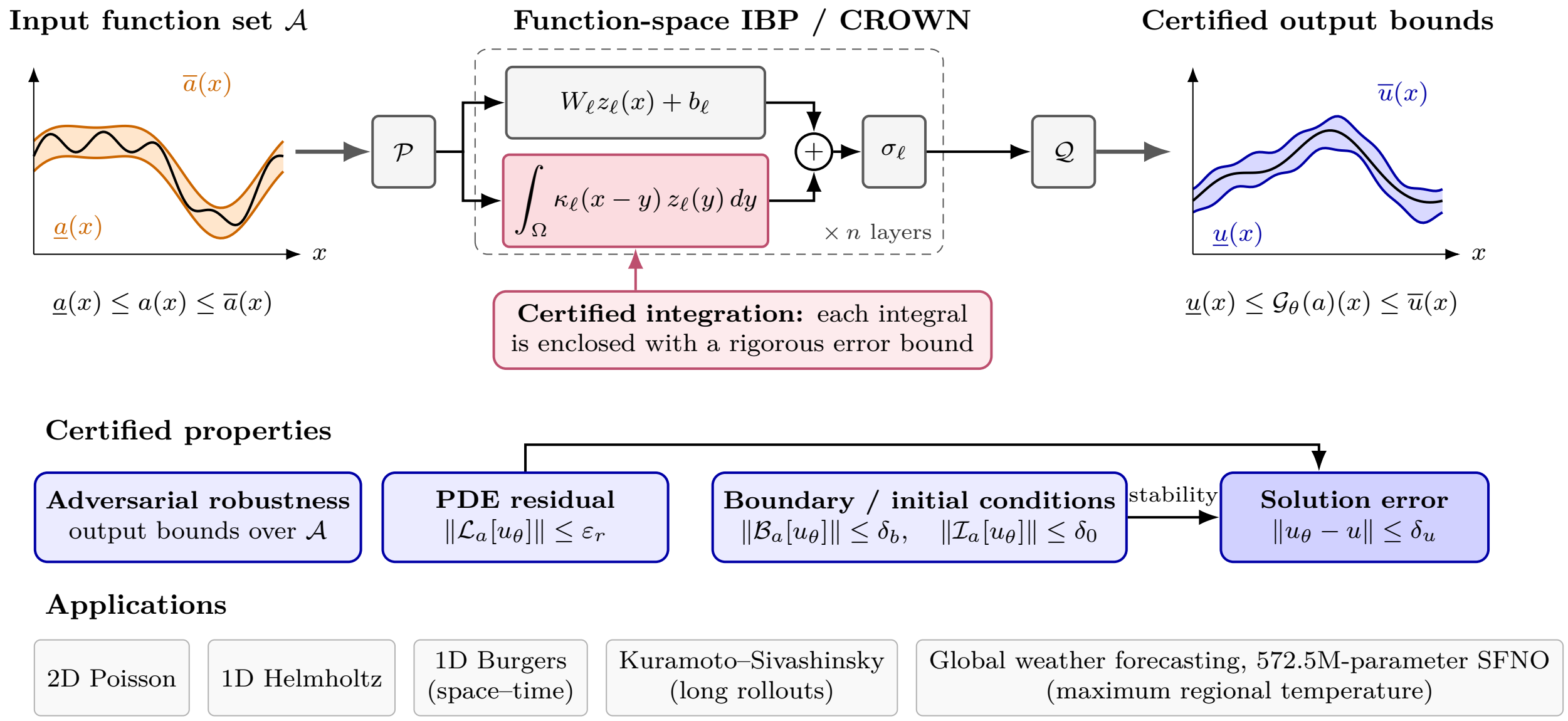

Machine learning models are often deployed in real-world applications where
test accuracy alone is insufficient, and properties such as safety, stability,
and robustness are critical factors in determining model reliability. In
scientific machine learning, reliability further requires physical
consistency, adherence to physical constraints, and control of worst-case
errors. In particular, data-driven approaches to solving PDEs~\citep{karniadakis2021physics, li2021fourierneuraloperatorparametric, Lu_2021} generally lack certified error bounds~\citep{eiras2024efficienterrorcertificationphysicsinformed} over the continuous domain, compared to traditional numerical methods~\citep{brenner2008fem, trefethen2000spectral} which offer convergence criteria and a posteriori error estimates~\citep{10.1093/acprof:oso/9780199679423.001.0001}. This has driven research in neural network \emph{certification}~\citep{katz2017reluplexefficientsmtsolver},
which provides provable guarantees that a desired property holds for every input in a prescribed set, rather than only for a finite collection of test examples.

A major class of certification methods is based on \emph{bound propagation}.
Interval bound propagation (IBP)~\citep{gowal2019effectivenessintervalboundpropagation} propagates lower and upper intervals through
the layers of a neural network, while CROWN~\citep{zhang2018efficient,xu2020automatic,xu2021fast,wang2021beta}
constructs tighter affine upper and lower bounds that retain dependencies
through the network. These algorithms establish \textit{adversarial robustness}, guaranteeing that model outputs remain
bounded over a set of inputs. Building on CROWN, $\partial$-CROWN~\citep{eiras2024efficienterrorcertificationphysicsinformed}
extends bound propagation to higher-order derivatives of physics-informed
neural networks (PINNs), allowing certification of
\begin{enumerate*}[label={(\arabic*)}]
\item initial conditions,
\item boundary conditions, and
\item PDE residual errors.
\end{enumerate*}

However, these methods are primarily developed for finite-dimensional neural networks.
Neural operators instead learn mappings between function spaces, for example,
mapping a forcing field to the solution field of a partial differential
equation (PDE). The Fourier neural operator
(FNO)~\citep{li2021fourierneuraloperatorparametric} is a widely used operator
learning architecture, while physics-informed neural operators
(PINOs)~\citep{li2023physicsinformedneuraloperatorlearning} can additionally
be trained with PDE residual losses to encourage physical consistency. For such models, certification should provide guarantees on the entire predicted function, rather than only on a finite-dimensional vector.

In practice, neural operators are implemented and evaluated on finite grids. Directly applying standard NN verification methods therefore certifies the model only at the grid points of the chosen resolution. Their certificate does not translate to arbitrary resolutions or to the continuum operator. When perturbations occur between grid points, grid-based verifiers may fail to capture their effects on the output function. This is distinct from discretization convergence: even when a discretized neural operator
converges to a continuum operator as the resolution increases, certification
at one fixed resolution does not by itself certify all admissible
functions consistent with those samples.

\begin{figure}[t]
  \centering
  \includegraphics[width=\linewidth]{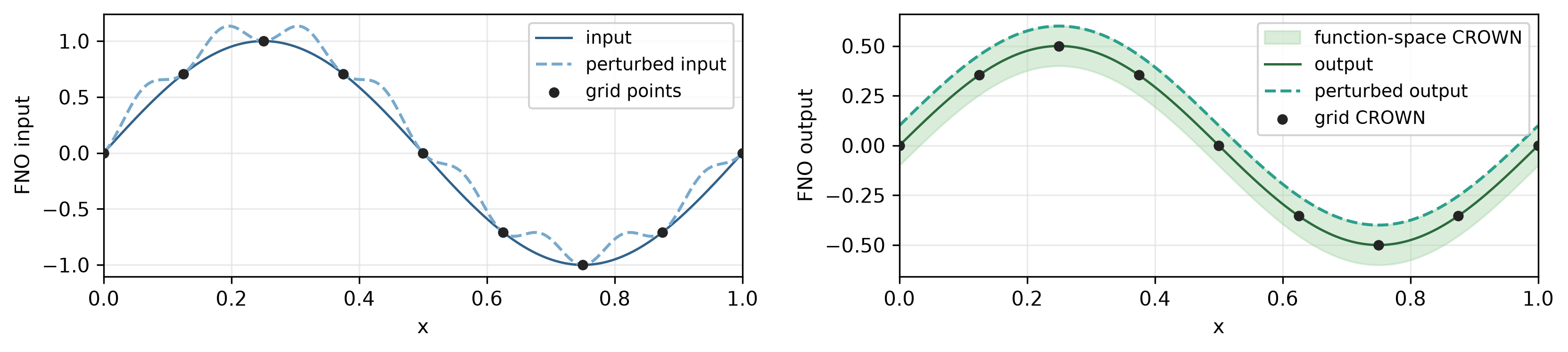}
  \caption{
  \textbf{Grid versus function-space certification counterexample.}
 Given two input functions that match at grid points, running CROWN on grid points (grid CROWN) can overlook perturbations between points, while function-space CROWN soundly bounds the outputs across the entire domain.
  }
  \label{fig:grid-not-continuum}
\end{figure}

We illustrate this with a simple counterexample: consider an input function $a(x)$ and perturbation $r(x)$ defined as
\[
a(x)=\sin(2\pi x),\qquad
r(x)=\frac{\varepsilon}{2}\left(1-\cos(2\pi N x)\right),
\]
over $[0,1)$ with $N=8$ and $\varepsilon=0.2$, and a neural operator $\mathcal G_\theta$ with Fourier multipliers
$R_0=1$, $R_{-1}=R_1=1/2$, and all others zero.
\begin{align*}
    \mathcal G_\theta(a)(x)
    &= \mathcal F^{-1}\!\left(R \cdot \mathcal F(a)\right)(x).
\end{align*}
where $\mathcal F$ denotes the Fourier transform. Then
\[
\mathcal G_\theta(a)(x)
=\tfrac12\sin(2\pi x),\qquad
\mathcal G_\theta(a_+)(x)
=\tfrac12\sin(2\pi x)+\tfrac{\varepsilon}{2}.
\]
However, at the sampled locations $x_j=j/N$, we have $r(x_j)=0$, so
$a(x)$ and $a_+(x)=a(x)+r(x)$ produce identical input vectors, thus grid-based verification observes zero input width at resolution $N$ even
though the continuum outputs differ everywhere. Figure~\ref{fig:grid-not-continuum}
illustrates this gap between grid and function-space certification.

\paragraph{Our approach.}
This motivates the study of neural operator certification in \textit{function space}, where
we formulate IBP and CROWN to efficiently propagate upper and lower bound functions for the
continuum FNO. Following the layer-by-layer propagation principle of standard IBP/CROWN, we derive propagation rules for each
component of the continuum FNO, including pointwise activations, lifting/projection layers, spectral convolutions, and normalization. Since our derived bounds are functions, they certify every point in the
domain for every function in the input set. Because spectral convolutions are integrals, we introduce a certified integration rule that computes integrals with rigorous error bounds using the Gaussian-Legendre quadrature~\citep{doi:10.1137/060659831,johansson2018numericalintegrationarbitraryprecisionball}.

We show that residual, boundary, and initial condition errors, which constitute the three PINN correctness conditions in $\partial$-CROWN~\citep{eiras2024efficienterrorcertificationphysicsinformed},
can be certified for the neural operator by extending it to higher-order
derivatives. For PDEs with suitable stability estimates, we further convert these errors into bounds on the neural operator's solution error~\citep{Hillebrecht_2025} without knowledge of the true PDE solution. This provides a post-training verification mechanism for learned surrogates over the continuous domain, analogous to the tolerance used to assess the accuracy of classical numerical solvers.

We evaluate our framework on neural operators of various sizes and architectures. Across the Poisson, Helmholtz, and Burgers' equations, we certify adversarial robustness, PDE residuals, and solution errors. To evaluate certification over long-horizon rollouts, we test our framework on the Kuramoto–Sivashinsky equation~\citep{wang2026coarsegrainingneuraloperators} for nonlinear chaotic dynamics. Finally, we scale our method to real-world global weather forecasting~\citep{Earth2Studio_Contributors_NVIDIA_Earth2Studio_2024}, certifying maximum regional temperature predictions for a 572.5-million-parameter pretrained spherical Fourier neural operator (SFNO)~\citep{bonev2023sphericalfourierneuraloperators}. The results show that our framework produces tight, domain-wide bounds across a diverse set of tasks while requiring no more than a few hours on a single GPU.

\section{Methods}
\subsection{Problem formulation}
We consider PDEs with boundary and initial conditions:
\begin{equation}
\begin{aligned}
    \mathcal L_a[u](x) &= 0,
    && x\in\Omega,\\
    \mathcal B_a[u](x) &= 0,
    && x\in\Gamma_b,\\
    \mathcal I_a[u](x) &= 0,
    && x\in\Gamma_0.
\end{aligned}
\label{eq:pde}
\end{equation}
Here $\Omega$ is the spatial or space--time domain, and
$\mathcal L_a$, $\mathcal B_a$, and $\mathcal I_a$ encode the governing
equation, boundary conditions, and initial conditions, respectively.
Let $\mathcal G:a\mapsto u$ denote the solution or forecasting operator.
The neural operator $\mathcal G_\theta$ approximates $\mathcal G$,
with $u_\theta=\mathcal G_\theta(a)$.

We define the continuum FNO as a function-to-function mapping through
$$
z_0(x)=\mathcal{P}(a(x)),\qquad
z_{\ell+1}(x)
=
\sigma_\ell\!\left(
W_\ell z_\ell(x)+b_\ell+
\int_\Omega \kappa_\ell(x-y)z_\ell(y)\,dy
\right),
$$

for \(\ell=0,\ldots,n-1\), where \(n\) is the number of FNO layers, followed by
$$
u(x)=\mathcal{Q}(z_n(x)).
$$
Here \(\mathcal{P}\) and \(\mathcal{Q}\) are pointwise lifting and projection maps, $\sigma_\ell$ is the pointwise activation, \(W_\ell\) is a local linear transformation, and the kernel is $\kappa_\ell(t)
=\frac1L\sum_{k=-(m-1)}^{m-1}
R_{\ell,k}\,e^{2\pi\mathrm{i}kt/L}$. At resolution $N$, the discrete FNO $\mathcal G_{\theta,N}$ replaces each integral by the sum
\[
    \frac{L}{N}\sum_{j=0}^{N-1}
    \kappa_\ell(x_i-x_j)z_\ell(x_j),
    \qquad x_j=jL/N,
\]
evaluated using the fast Fourier transform (FFT).

Following prior NN verification work~\citep{katz2017reluplexefficientsmtsolver,gowal2019effectivenessintervalboundpropagation,zhang2018efficient} and $\partial$-CROWN~\citep{eiras2024efficienterrorcertificationphysicsinformed}, our framework verifies the following conditions for a neural operator $\mathcal G_\theta$:

\begin{enumerate}[leftmargin=*]
    \item \textbf{Regional maxima.}
    For a predicted field $u_\theta=\mathcal G_\theta(a)$ over a region $R$, we certify
    \[
        \max_{x\in R}u_\theta(x) \le u_{\text{max}},
    \]

    \item \textbf{Adversarial robustness.}
    We consider a smooth input set \(\mathcal A\subset\mathcal X\),
$$
\mathcal A
=
\left\{
a \;\middle|\;
\underline a(x)\le a(x)\le\overline a(x)
\,\,\,\forall x\in\Omega,\quad |a'(x)|\le M_a
\right\}.
$$
We construct upper/lower bound functions satisfying
    \[
        \underline u(x)\le u_\theta(x)\le\overline u(x)
        \qquad\forall a\in\mathcal A,\quad\forall x\in\Omega.
    \]
    The formal theorem statement and proof are in
    Appendix~\ref{app:soundness}.

    \item \textbf{PDE error.}
    For a fixed input $a$, we bound the PDE residual over the entire domain:
\begin{align}\label{res}
    \|\mathcal L_a[\mathcal G_\theta(a)]\|
    \le \varepsilon_r,
    \qquad
\end{align}
In addition, we bound the boundary and initial
condition errors as
\begin{equation}\label{boundary}
\begin{aligned}
    \|\mathcal B_a[\mathcal G_\theta(a)]\|
    &\le \delta_b,\quad
    \|\mathcal I_a[\mathcal G_\theta(a)]\|
    \le \delta_0.
\end{aligned}
\end{equation}
\item \textbf{Solution error.}
Combining residual and boundary/initial conditions, we bound the error relative to the exact solution $u=\mathcal G(a)$:
\begin{equation}\label{solution-error}
    \|\mathcal G_\theta(a)-u\|\le\delta_u.
\end{equation}
\end{enumerate}

\subsection{Certified integration}\label{sec:certified-integration}
Each spectral convolution layer integrates a hidden function \(z_\ell(y)\)
against a smooth kernel \(\kappa_\ell(x-y)\). Standard numerical evaluation simply approximates this by a weighted sum. For rigorous certification, we
instead bound the approximation error explicitly so that the true integral is
guaranteed to lie within our lower and upper bounds. Inspired by Arb~\citep{johansson2018numericalintegrationarbitraryprecisionball}, we use Gaussian quadrature, and choose its error bound depending on whether the integrand is smooth.

\begin{itemize}[leftmargin=*]
\item \textbf{Smooth case.}
When the integrand is \emph{analytic}, as for $\tanh$ networks, we partition the domain into cells \(C_j=[x_j,x_j+h)\) of width \(h=L/Q\) and use the Gauss--Legendre quadrature~\citep{doi:10.1137/060659831, Trefethen:2011:SMP}. It chooses $n$ points $x_i$ and weights $w_i$,
\[
    \int_{-1}^{1}f(x)\,dx \approx \sum_{i=1}^{n} w_i f(x_i),
\]
such that the formula integrates all polynomials of degree up to \(2n-1\) exactly. In this case, $x_i$ is chosen to be the roots of the Legendre polynomial $P_{n}(x)$. The error is bounded by a Cauchy remainder that decays as \(4^{-2n}\). We define \(Q\) as the \textit{quadrature resolution}.

\item \textbf{Non-smooth case.}
In the case where the integrand is not analytic, for example when the model uses ReLU activations or when the input is perturbed, the Cauchy remainder no longer decays exponentially. We instead bound the error using only the slope of the integrand. We apply the two-point Gauss--Legendre rule to each cell, with nodes \(x_{j,k}\) and weights \(h/2\). Suppose we are given \(\underline f_{j,k}\), \(\overline f_{j,k}\), \(M_{f,j}\) such that
\[
    \underline f_{j,k}\le f(x_{j,k})\le\overline f_{j,k},
    \qquad
    \sup_{s\in C_j}|f'(s)|\le M_{f,j}.
\]
Then the continuum integral satisfies
\begin{equation}
    \frac{h}{2}\sum_{j,k}\underline f_{j,k}-\eta_f
    \le \int_\Omega f(s)\,ds
    \le \frac{h}{2}\sum_{j,k}\overline f_{j,k}+\eta_f,
    \qquad
    \eta_f=d\,h^2\sum_{j=0}^{Q-1}M_{f,j},
    \label{eq:corrected-quadrature-proposition}
\end{equation}
where \(d=5/12-\sqrt3/6\approx0.128\) is the average distance from a point of the cell to its nearest node, in units of \(h\). The integration error decreases as \(\mathcal{O}(Q^{-1})\).

\end{itemize}

More details on the error bounds and quadrature methods are discussed in Appendix~\ref{app:crown-quadrature}. We next use this integration rule within function-space IBP and CROWN.
\subsection{Interval bound propagation}\label{sec:ibp}
Function-space IBP propagates both nodal value intervals and cellwise
derivative intervals. For every hidden field \(z\) and cell \(C_j\), we
maintain
\begin{equation}
    \underline z(x_j)\le z(x_j)\le\overline z(x_j),
    \qquad
    \underline z'_j\le z'(x)\le\overline z'_j
    \quad\text{for every }x\in C_j.
    \label{eq:ibp-maintained-state}
\end{equation}
Pointwise affine maps and activations use standard IBP
and chain-rule enclosures, the update rules are detailed in
Appendix~\ref{app:ibp-rules}. For the spectral
convolution layer,
\[
    y(x)=b+\int_\Omega \kappa(x-s)z(s)\,ds.
\]
Sign splitting gives the continuum enclosure
\begin{align*}
    \overline y(x)
    &=
    b+\int_\Omega
        \kappa^+(x-s)\overline z(s)
        +\kappa^-(x-s)\underline z(s)
    ds,\\
    \underline y(x)
    &=
    b+\int_\Omega
        \kappa^+(x-s)\underline z(s)
        +\kappa^-(x-s)\overline z(s)
    ds.
\end{align*}
Applying the same rule to the derivative bound,
\begin{equation}
\begin{aligned}
    \overline y'(x)
    &=\int_\Omega
      \kappa'^+(x-s)\overline z(s)
      +\kappa'^-(x-s)\underline z(s)ds,\\
    \underline y'(x)
    &=\int_\Omega
      \kappa'^+(x-s)\underline z(s)
      +\kappa'^-(x-s)\overline z(s)ds.
\end{aligned}
\label{eq:ibp-spectral-derivative}
\end{equation}
Each integral is enclosed using the certified integration rules in Section~\ref{sec:certified-integration}. After the final layer, suppose
\(\underline y_j\leq y(x_j)\leq\overline y_j\) at the left node of
cell \(C_j=[x_j,x_j+h)\). Then, for every
\(x\in C_j\),
\begin{align}\label{extend}
    \underline y_j-M_{y,j}(x-x_j)
    \leq y(x)\leq
    \overline y_j+M_{y,j}(x-x_j).
\end{align}
This certifies every point between nodes, not only the quadrature grid. For two-dimensional domains, we tighten this forward propagation by combining bilinear enclosures with linearization, as described in Appendix~\ref{app:2d-linearization}.
\subsection{CROWN}\label{sec:crown}
When verifying FNOs using standard CROWN, we propagate bounds in the following form:
$$\mathcal{B}_{x_i}[z] = d_i + {p}_i^\top {z}(x_i) + \sum_{j=0}^{N-1} {c}_{ij}^\top {z}(x_j).$$
Function-space CROWN instead maintains lower and upper affine \textit{functionals},
$$
    \mathcal B_x[z]=d(x)+p(x)^\top z(x)
    +\int_\Omega c(x,s)^\top z(s)\,ds.
$$
Here \(d\) is the bias, \(p\) is the pointwise coefficient, and \(c\) is the
kernel coefficient. Unlike standard CROWN's finite-dimensional affine forms, \(B_x\) contains both a local coefficient \(p(x)\) and a kernel coefficient \(c(x,\cdot)\), allowing the bound to retain dependence on the entire input function. Backpropagation starts at \(d=0\), \(p=1\), and \(c=0\).
Pointwise affine maps and activation relaxations use the usual CROWN
substitutions. For the spectral convolution layer \(y(s)=Wz(s)+b+\int_\Omega K(s-t)z(t)\,dt\), the update rule is given by
\begin{align*}
    \widetilde d(x)&=d(x)+p(x)^\top b
    +\int_\Omega c(x,s)^\top b\,ds,\\
    \widetilde p(x)&=W^\top p(x),\\
    \widetilde c(x,t)&=W^\top c(x,t)+K(x-t)^\top p(x)
    +\int_\Omega K(s-t)^\top c(x,s)\,ds.
\end{align*}
After backpropagating to the input, let \((d^L,p^L,c^L)\) and \((d^U,p^U,c^U)\) denote the lower and
upper affine functionals. Concretizing them over the input box gives the
output bound functions
\begin{align*}
    \underline u(x)
    =d^L(x)+(p^L)^+(x)^\top\underline a(x)
      +(p^L)^-(x)^\top\overline a(x)
    +\int_\Omega\!
      (c^L)^+(x,s)^\top\underline a(s)
      +(c^L)^-(x,s)^\top\overline a(s)ds,\\
    \overline u(x)
    =d^U(x)+(p^U)^+(x)^\top\overline a(x)
      +(p^U)^-(x)^\top\underline a(x)
    +\int_\Omega\!
      (c^U)^+(x,s)^\top\overline a(s)
      +(c^U)^-(x,s)^\top\underline a(s)ds,
    \label{eq:crown-bound-functions}
\end{align*}
Thus \(\underline u(x)\leq u_\theta(x)\leq\overline u(x)\)
for every \(a\in\mathcal A\), where \(u_\theta=\mathcal G_\theta(a)\). Appendix~\ref{app:crown-derivation}
derives all affine, activation, spectral, derivative, and input-concretization rules, while Appendix~\ref{app:crown-quadrature} gives their validated
quadrature remainders. After certifying each node, we extend to points in between nodes using the same idea as IBP in \eqref{extend}, thus providing enclosures for every point in the domain.

Function-space IBP on a $d$-dimensional grid with $Q$ points per axis costs $\mathcal{O}(Q^d\log Q)$, while function-space CROWN propagates a separate backward pass for each of the $Q^d$ output cells and therefore scales as $\mathcal{O}(Q^{2d}\log Q)$ due to the underlying FFTs.

\subsection{Residual Certification}\label{sec:residual-certification}
Robustness certification controls how the FNO output varies under input perturbations, but it does not establish that a prediction obeys the governing physical laws. Classical numerical solvers verify accuracy on each solve using explicit stopping tolerances and residual checks~\citep{saad2003iterative,hairer1993solving}, whereas learned PDE surrogates produce forward predictions in a single pass without built-in stopping criteria or per-instance tolerance checks~\citep{grossmann2024can,mcgreivy2024weak}. Because the exact solution $u=\mathcal G(a)$ is unknown at inference time, we cannot directly evaluate the error $u_\theta-u$. Instead, we certify the PDE residual and boundary or initial conditions over the continuous domain.

We consider the same verification problem as $\partial$-CROWN~\citep{eiras2024efficienterrorcertificationphysicsinformed}, where we fix the forcing term and certify the worst-case residual and boundary error of the PDE. However, our verification methodology is fundamentally different from that of $\partial$-CROWN. A PINN is designed to predict the solution field $u_\theta(x,t)$ from space--time coordinates $(x,t)$, so $\partial$-CROWN treats the domain as an input box for bound propagation with CROWN. In contrast, the FNO directly maps the forcing function $a$ to the solution field $u_\theta$. Since our input is a fixed forcing function, we do not propagate input perturbations and can therefore use fixed-input IBP rather than CROWN to obtain tight residual bounds.

Certifying residual errors requires bounds on higher-order partial derivatives of the solution field. For instance, the residual of the Burgers equation is
$$
    r =u_t+uu_x-\nu u_{xx}.
$$
Given a fixed input $a$, we extend IBP to include second derivatives. For spectral layers,
$$
    y^{\prime\prime}(x)=\int_\Omega\kappa^{\prime\prime}(x-s)z(s)\,ds.
$$
For the activation layers, we use $\tanh$ instead of ReLU, whose classical second derivative is undefined at the origin:
$$
    y^\prime=\sigma^\prime(z)z^\prime,\qquad
    y^{\prime\prime}=\sigma^{\prime\prime}(z)(z^\prime)^2+\sigma^\prime(z)z^{\prime\prime}.
$$
In general, we support PDE residuals with derivatives up to fourth order.

\subsection{Solution Certification}\label{sec:solution-certification}
While residual certification verifies how closely $u_\theta$ satisfies the differential equation locally, a small PDE residual does not by itself guarantee a small solution error $\|u_\theta-u\|$~\citep{mishra2023estimates,Hillebrecht_2025}. When the differential operator is ill-conditioned or near resonance, or when errors accumulate over time in nonlinear evolution equations, a modest residual or boundary mismatch can still produce a large deviation from the exact solution $u=\mathcal G(a)$.

To bridge the gap from residual bounds to guaranteed solution accuracy---analogous to \emph{a posteriori} error estimators in adaptive finite element and spectral methods~\citep{babuska1978error,ainsworth2000posteriori,10.1093/acprof:oso/9780199679423.001.0001} and validated numerics~\citep{nakao2019numerical}---we follow \emph{a posteriori} PDE error analysis~\citep{mishra2023estimates,deryck2022generic,Hillebrecht_2025,ernst2025certified} and combine our residual bound and boundary/initial condition bounds with stability estimates for the governing PDE~\citep{evans2010partial,gilbarg2001elliptic} to bound the true solution error. These estimates are PDE-specific, for example~\citep{mishra2023estimates},
\[
\delta_u=C_r\varepsilon_r+C_b\delta_b+C_0\delta_0,
\]
where $C_r,C_b,C_0\ge 0$ are stability constants that depend on the differential operator and domain geometry. Substituting our certified bounds $\varepsilon_r,\delta_b,\delta_0$ directly yields the guaranteed upper bound $\delta_u$ without requiring the true solution $u$.

To derive this bound for a given PDE, one substitutes the error $e=u_\theta-u$ into the PDE so that $e$ satisfies a differential equation driven by the interior residual $\mathcal L_a[u_\theta]$ and boundary or initial mismatches $\mathcal B_a[u_\theta],\mathcal I_a[u_\theta]$~\citep{mishra2023estimates}. For linear PDEs, bounding the inverse of the differential operator via the maximum principle or Green's function estimates~\citep{gilbarg2001elliptic,evans2010partial} directly determines $C_r$ and $C_b$. For example, for the Poisson equation $- u_{xx}=a$ on $[0,L]$ with Dirichlet boundary conditions, applying the maximum principle gives $C_r=L^2/8$, $C_b=1$, $C_0=0$ in the $L^\infty$ norm:
\[
\|u_\theta-u\|_\infty  \le \delta_u =\frac{L^2}{8}\varepsilon_r + \delta_b.
\]
Appendix~\ref{app:solution-stability} derives the explicit stability estimates for each PDE studied in our experiments.

\section{Experiments}
We evaluate our certification framework across six neural operator benchmarks and four verification problems, as detailed in the following sections.
\subsection{PDE overview}
We consider the following PDEs for robustness and solution-error certification:
\begin{enumerate}[leftmargin=*]
    \item \textbf{Poisson equation}
The Poisson equation models how a forcing function drives the potential field, frequently used in electrostatics, gravity, and fluid dynamics. Its PDE and learning problem are
\[
-\nabla^2 u(x) = a(x),\quad x\in\Omega,\qquad \mathcal{G}_\theta:a\mapsto u_\theta,
\]
with boundary conditions $u=0$ on $\partial\Omega$, where $\Omega=(0,1)$ in 1D and $\Omega=(0,1)^2$ in 2D. Its linear and smooth behavior indicates it is comparatively easier to train. We train a 1D FNO with one Fourier layer, eight hidden channels, 24 Fourier modes, and grid positional embedding at resolution 64, and a 2D FNO with two Fourier layers, eight hidden channels, and $12\times 12$ Fourier modes at resolution $32\times 32$.

    \item \textbf{Helmholtz equation}
To stress-test our framework, we choose the Helmholtz equation,
\[
-\frac{\partial^2u(x)}{\partial x^2}-\kappa^2u(x) = a(x),\quad x\in[0,1),\qquad \mathcal{G}_\theta:a\mapsto u_\theta,
\]
with boundary conditions $u(0)=u(1), \,\,u'(0)=u'(1)$. It is resonant when the frequency approaches $\kappa$ with gain $\frac{1}{\omega^2-\kappa^2}\gg 1$. This resonance violently amplifies any input perturbations and subsequently expands the certified bounds. We train an FNO with one Fourier layer, four hidden channels, 32 Fourier modes, and no positional embedding at resolution 64.

    \item \textbf{Burgers equation}
We also choose the Burgers equation, which is a nonlinear PDE for viscous fluid flow and one of the main PDEs used by~\citep{li2021fourierneuraloperatorparametric},
\begin{align*}
&\frac{\partial u(x,t)}{\partial t}
+u(x,t)\frac{\partial u(x,t)}{\partial x}
=
\nu\frac{\partial^2u(x,t)}{\partial x^2}, \quad
x\in[0,1),\ t\in(0,1],\qquad
\mathcal{G}_\theta:\ u(x,0)\mapsto u_\theta(x,t),
\end{align*}
where $\nu=0.01$ and the boundary conditions are $u(0,t)=u(1,t)$ and $\partial_xu(0,t)=\partial_xu(1,t)$. We train a 2D FNO with 2 Fourier layers, 16 hidden channels, and $16\times16$ Fourier modes.
\end{enumerate}
\subsection{Adversarial robustness}
\begin{figure}
    \centering
    \includegraphics[width=1\linewidth]{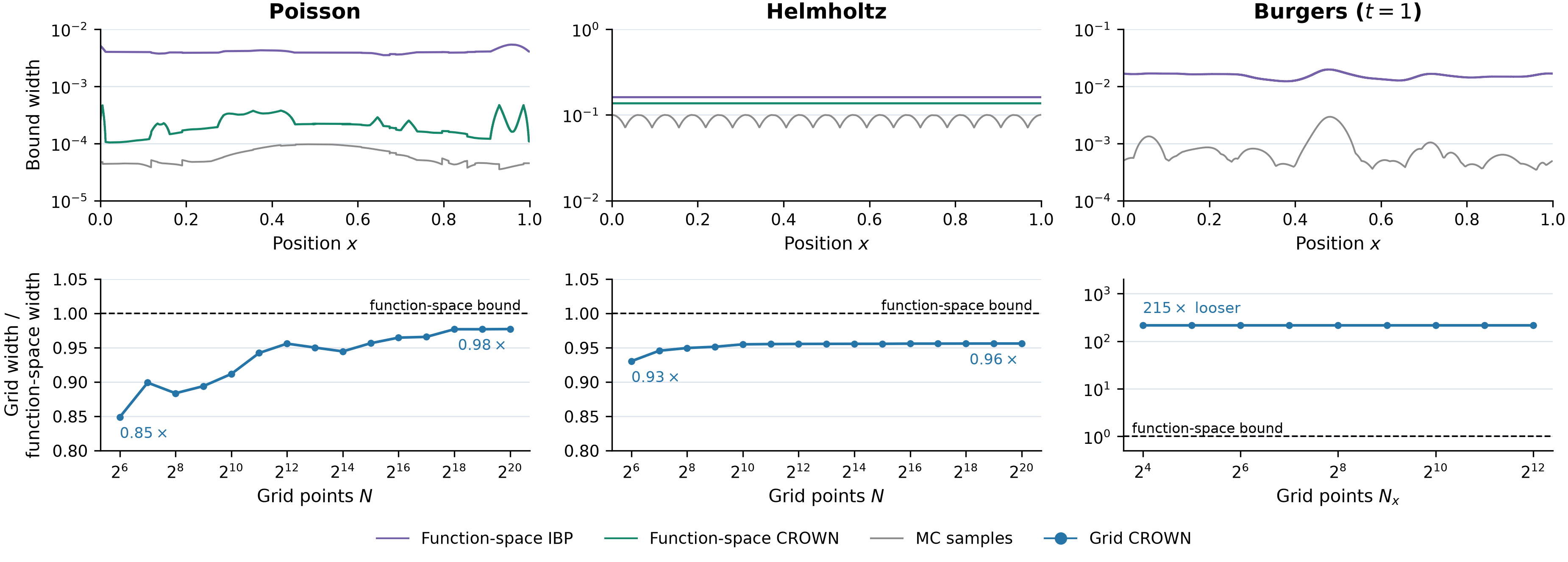}
\caption{
\textbf{Grid certificates cover only grid points, function-space certificates cover every position.}
\\\textbf{Top:} Width of the output bounds at each position $x$ for input perturbations of size $\varepsilon=10^{-3}$, from function-space IBP (purple) and function-space CROWN (green). Gray shows MC sample widths. Smaller bound widths are tighter. Burgers is shown at the final time $t=1$.
\\\textbf{Bottom:} Maximum grid CROWN width across different resolutions $N$, divided by the function-space width (dashed line). Overall, function-space bounds are comparable to or tighter than grid bounds.}
    \label{fig:flat-quad}
\end{figure}
We evaluate the tightness and cost of function-space IBP and CROWN on continuum FNOs trained for Poisson, Helmholtz, and Burgers. Specifically, we use the max bound width established in Theorem~\ref{thm:function-soundness} as our evaluation metric, $
    \max_j \big(\overline u(x_j)-\underline{u}(x_j) + 2M_{u,j}h\big).
$ For 2D FNOs, CROWN is intractable due to the $\mathcal{O}(Q^4\log Q)$ time complexity, while for IBP we use a tighter variant described in Appendix~\ref{app:2d-linearization}.

For reference, we include grid-level IBP and CROWN using Auto-LiRPA~\citep{xu2020automatic} on the discretized FNO at the training resolution. (Note that the grid and function-space methods certify different operators.) We compare them with empirical widths obtained by Monte Carlo sampling~\citep{metropolis1949monte}. The quadrature resolution is \(Q =1{,}048{,}576\). We treat $Q$ as a tunable hyperparameter and choose it to be as large as possible within our computational budget. Full-domain grid and quadrature experiments are conducted on a single NVIDIA B300 GPU. We benchmark running time, peak memory usage (GiB), bound width, and gap, defined as the ratio against the empirical width. The complete experimental configurations are provided in Appendix~\ref{app:experiment-details}.
\begin{table}
\centering
\footnotesize
\caption{
Robustness certification for grid and function-space IBP/CROWN on an NVIDIA B300 GPU,
using Gaussian quadrature with \(Q=1{,}048{,}576\).
Function-space IBP uses linearized perturbation bounds at \(2048\times2048\) for 2D Poisson and \(4096\times4096\) for space--time Burgers.
}
\label{tab:main-results}
\begin{tabular}{llllrrrrr}
\toprule

& PDE
& Scope
& Method
& Bound width
& Empirical width
& Gap
& Time (s)
& Memory (GiB)\\
\midrule

1D
& Poisson
& Grid
& IBP
& \(5.330\times10^{-3}\)
& \(9.771\times10^{-5}\)
& \(54.54\)
& \(0.08\)
& \(0.008\)\\

&&
& CROWN
& \(3.981\times10^{-4}\)
& \(9.771\times10^{-5}\)
& \(4.07\)
& \(0.36\)
& \(0.010\)\\

&& Function-space
& IBP
& \(5.364\times10^{-3}\)
& \(9.580\times10^{-5}\)
& \(55.98\)
& \(0.55\)
& \(3.844\)\\

&&
& CROWN
& \(4.689\times10^{-4}\)
& \(9.580\times10^{-5}\)
& \(4.89\)
& \(1833.87\)
& \(18.699\)\\

& Helmholtz
& Grid
& IBP
& \(1.593\times10^{-1}\)
& \(1.000\times10^{-1}\)
& \(1.59\)
& \(0.03\)
& \(0.008\)\\

&&
& CROWN
& \(1.273\times10^{-1}\)
& \(1.000\times10^{-1}\)
& \(1.27\)
& \(0.15\)
& \(0.009\)\\

&& Function-space
& IBP
& \(1.609\times10^{-1}\)
& \(9.972\times10^{-2}\)
& \(1.61\)
& \(0.08\)
& \(1.173\)\\

&&
& CROWN
& \(1.368\times10^{-1}\)
& \(9.972\times10^{-2}\)
& \(1.37\)
& \(243.43\)
& \(9.122\)\\

\midrule
2D
& Poisson
& Grid
& IBP
& \(3.211\times10^{-1}\)
& \(9.341\times10^{-5}\)
& \(3436.37\)
& \(0.05\)
& \(0.012\)\\

&&
& CROWN
& \(4.217\times10^{-2}\)
& \(9.341\times10^{-5}\)
& \(451.39\)
& \(1.88\)
& \(0.058\)\\

&& Function-space
& IBP
& \(4.646\times10^{-4}\)
& \(6.163\times10^{-5}\)
& \(7.54\)
& \(9.84\)
& \(148.043\)\\

& Burgers
& Grid
& IBP
& \(5.275\)
& \(5.299\times10^{-3}\)
& \(995.12\)
& \(0.03\)
& \(0.015\)\\

&&
& CROWN
& \(4.237\)
& \(5.299\times10^{-3}\)
& \(799.39\)
& \(0.31\)
& \(0.026\)\\

&& Function-space
& IBP
& \(2.466\times10^{-2}\)
& \(5.072\times10^{-3}\)
& \(4.86\)
& \(121.31\)
& \(126.249\)\\
\bottomrule
\end{tabular}
\end{table}


The results are shown in Table~\ref{tab:main-results} and Figure~\ref{fig:flat-quad}. The best empirical gaps are consistently below \(10\times\) across the four PDEs, with bound widths comparable to or tighter than those obtained by grid IBP/CROWN across different resolutions. In particular, our linearized IBP method significantly outperforms grid IBP/CROWN on the 2D PDEs, showing the scalability and tightness of our approach. Furthermore, function-space bounds also successfully bound the resonant behavior of the Helmholtz equation, achieving a gap of only \(1.37\times\). All certification runs can be completed within 1 hour and uses at most 150 GiB of memory.

\subsection{PDE residual and solution-error certification}

We next certify how closely each trained FNO satisfies its governing PDE over the continuous domain. Because residual evaluation requires second-order spatial derivatives, we evaluate FNOs with smooth $\tanh$ activations for a fixed input using function-space IBP and Gaussian quadrature. We partition the domain into $2^{25}$ cells for 1D Poisson and Helmholtz, $16384\times 16384$ cells for 2D Poisson, and $16384\times 4096$ space--time cells for Burgers. In our implementation, a single certification pass simultaneously yields both $L_\infty$ and $L_2$ residual bounds.

Table~\ref{tab:residual-results}(a) reports the certified upper bound on the residual error along with the bound width (the difference between the certified upper and lower bounds). The true residual is guaranteed to lie within this interval of width of order $10^{-3}$ across all cases, confirming that our certified bounds are tight. Every run finishes in under 10 minutes using at most $3.23$ GiB of GPU memory. The Helmholtz model has a larger residual than 1D Poisson (\(0.283\) versus \(0.111\)) because near-resonance amplifies small prediction errors.
\begin{table}[t]
\centering
\footnotesize
\caption{Residual and solution error certification. (a) Certified residual bounds and their widths. (b) Certified solution error bounds from PDE stability estimates. In (b), bound width measures the uncertainty in the stability expression, the true solution error may be smaller than the lower bound.}
\label{tab:residual-results}
\label{tab:solution-results}
\setlength{\tabcolsep}{3pt}
\begin{tabular*}{\linewidth}{@{\extracolsep{\fill}}lllrrrrr@{}}
\multicolumn{8}{@{}l}{\textbf{(a) PDE residual bounds}}\\[4pt]
\toprule
& PDE
& \multicolumn{2}{l}{Quantity}
& Upper bound
& Bound width
& Time (s)
& Memory (GiB)\\
\midrule
1D
& Poisson
& \multicolumn{2}{l}{\(\|{-\partial_{xx}u_\theta-a}\|_\infty\)}
& \(1.105\times10^{-1}\)
& \(1.061\times10^{-5}\)
& \(29.11\)
& \(2.013\)\\
&& \multicolumn{2}{l}{\(\|{-\partial_{xx}u_\theta-a}\|_2\)}
& \(5.515\times10^{-2}\)
& \(1.451\times10^{-5}\)
&
& \\
& Helmholtz
& \multicolumn{2}{l}{\(\|{-\partial_{xx}u_\theta-\kappa^2u_\theta-a}\|_\infty\)}
& \(2.833\times10^{-1}\)
& \(5.875\times10^{-4}\)
& \(37.67\)
& \(2.013\)\\
&& \multicolumn{2}{l}{\(\|{-\partial_{xx}u_\theta-\kappa^2u_\theta-a}\|_2\)}
& \(1.177\times10^{-1}\)
& \(6.864\times10^{-4}\)
&
& \\
\midrule
2D
& Poisson
& \multicolumn{2}{l}{\(\|{-\partial_{xx}u_\theta-\partial_{yy}u_\theta-a}\|_\infty\)}
& \(2.935\)
& \(1.124\times10^{-3}\)
& \(533.43\)
& \(3.223\)\\
&& \multicolumn{2}{l}{\(\|{-\partial_{xx}u_\theta-\partial_{yy}u_\theta-a}\|_2\)}
& \(8.491\times10^{-1}\)
& \(1.439\times10^{-3}\)
&
& \\
& Burgers
& \multicolumn{2}{l}{\(\|{\partial_{t}u_\theta+u_\theta\partial_{x}u_\theta-\nu\partial_{xx}u_\theta}\|_{2}\)}
& \(2.703\times10^{-1}\)
& \(3.539\times10^{-3}\)
& \(378.15\)
& \(1.578\)\\
\bottomrule
\addlinespace[8pt]
\multicolumn{8}{@{}l}{\textbf{(b) Solution error bounds}}\\[4pt]
\toprule
    & PDE
    & Quantity
    & Reference error
    & Upper bound
    & Bound width
    & Time (s)
    & Memory (GiB)\\
    \midrule
    1D
    & Poisson
    & \(\|u_\theta - u\|_\infty\)
    & \(3.113\times10^{-3}\)
    & \(1.623\times10^{-2}\)
    & \(1.325\times10^{-6}\)
    & \(29.11\)
    & \(2.013\)\\
    && \(\|u_\theta - u\|_2\)
    & \(2.331\times10^{-3}\)
    & \(8.011\times10^{-3}\)
    & \(1.469\times10^{-6}\)
    &
    & \\
    & Helmholtz
    & \(\|u_\theta - u\|_\infty\)
    & \(6.989\times10^{-4}\)
    & \(18.03\)
    & \(3.740\times10^{-2}\)
    & \(37.67\)
    & \(2.013\)\\
    && \(\|u_\theta - u\|_2\)
    & \(2.757\times10^{-4}\)
    & \(5.882\)
    & \(3.432\times10^{-2}\)
    &
    & \\
    \midrule
    2D
    & Poisson
    & \(\|u_\theta - u\|_\infty\)
    & \(4.910\times10^{-3}\)
    & \(3.718\times10^{-1}\)
    & \(1.410\times10^{-4}\)
    & \(533.43\)
    & \(3.223\)\\
    && \(\|u_\theta - u\|_2\)
    & \(2.054\times10^{-3}\)
    & \(4.799\times10^{-2}\)
    & \(7.300\times10^{-5}\)
    &
    & \\
& Burgers
& \(\|u_\theta - u\|_{2}\)
& \(2.368\times10^{-3}\)
& \(1.023\)
& \(5.862\times10^{-2}\)
& \(378.15\)
& \(1.578\)\\
\bottomrule
\end{tabular*}
\end{table}

Combining the certified residual bounds with boundary and initial condition errors yields a guaranteed upper bound on the true solution error $\|u_\theta - u\|$ without requiring the exact PDE solution (see Appendix~\ref{app:solution-stability} for the conversion formulas). Table~\ref{tab:solution-results}(b) reports the certified upper bounds and bound width, along with the reference error obtained by comparing with the exact PDE solution. 

For residual bounds, these indicate that the maximum residual over the domain is within the interval $[\text{Upper bound} - \text{Bound width}, \text{Upper bound}]$. For solution error bounds, we only certify that the solution estimates lie within the interval, the true solution error may be smaller than the lower bound. For 1D Poisson, the upper bound closely tracks the reference error ($8.011\times 10^{-3}$ versus $2.331\times 10^{-3}$ in $L^2$). For Helmholtz, 2D Poisson, and Burgers, however, the certified upper bounds are substantially looser than the reference errors. This gap stems from the PDE stability analysis rather than looseness in our residual certification. Since the stability formulas assume worst-case conditions, they can be conservative in their estimations.

\subsection{Multi-step rollout certification}
To further investigate the scalability of framework, we certify long-term dynamics of chaotic systems, which is difficult to predict due to its sensitivity to small perturbations. Following~\citep{li2022learningdissipativedynamicschaotic, wang2026coarsegrainingneuraloperators}, we approximate long-term dynamics through autoregressive rollouts of the neural operator, 
\begin{align*}
    u(n\tau) \approx \underbrace{(\mathcal{G}_\theta \circ \cdots \circ \mathcal{G}_\theta)}_{n\text{ times}}(u_0),\quad \mathcal{G}_\theta:\ u(t)\mapsto u_\theta(t+\tau).
\end{align*}
We consider the Kuramoto–Sivashinsky (KS) equation, a well-studied PDE for modeling spatiotemporal chaos, 
\begin{align*}
    \frac{\partial u(x,t)}{\partial t} + u(x,t) \frac{\partial u(x,t)}{\partial x} + \frac{\partial^2 u(x,t)}{\partial x^2} + \nu \frac{\partial^4 u(x,t)}{\partial x^4} = 0,\quad x\in[0,6\pi),\quad t\in(0,\infty).
\end{align*}
with periodic boundary conditions. The KS equation includes a fourth-order derivative, which makes certification particularly challenging. We train a PINO~\citep{li2023physicsinformedneuraloperatorlearning} with 1 layer and 64 channels, and perform rollout for 25 time steps with a step size $\tau=0.1$. For this experiment, we certify the residual bound at each time step $n\tau$.

\begin{figure}[t]
    \centering
    \includegraphics[width=\linewidth]{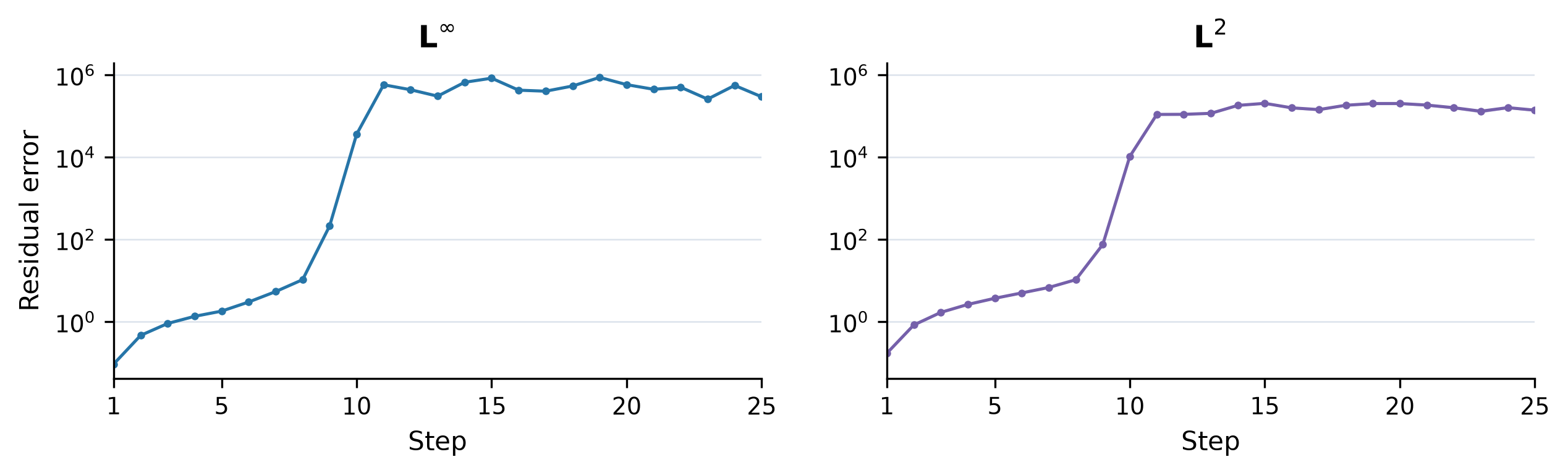}
    \caption{\textbf{Residual bounds across 25 autoregressive rollout steps of the Kuramoto--Sivashinsky equation.} We evaluate a neural operator trained on the KS equation with step size $\tau=0.1$ for 25 time steps. Shaded bands indicate certified intervals for spatial $L^\infty$ (left) and $L^2$ (right) residuals. The predicted field develops sharp spatial features around steps 9--11, causing the residual to jump from $\sim 10^1$ to $\sim 10^5$. This massive residual indicates that the model has failed to maintain physical consistency under long-horizon rollout, a failure mode our certified bounds expose.}
    \label{fig:ks-rollout}
\end{figure}

Figure~\ref{fig:ks-rollout} shows the certified residual bounds across all 25 rollout steps. During the first eight steps ($t \le 0.8$), the model produces physically consistent predictions, with certified residuals staying below $10$. By step 9–11, repeated feedback causes the predicted field to develop sharp spatial features. Because the KS residual includes derivatives up to fourth order, those features produce enormous residuals even when the field's amplitude looks moderate. Importantly, the certified bound widths (the shaded intervals) remain narrow across all 25 steps, with relative gaps between upper and lower bounds staying small ($<15\%$ for $L^2$ and $<4\%$ for $L^\infty$) throughout the rollout. This confirms that the sharp rise reflects real physical divergence in the model rather than looseness in the verifier. 
\subsection{Global weather forecast certification}
\noindent\begin{minipage}[t]{0.44\textwidth}
    \vspace{0pt}
    \centering
    \includegraphics[width=0.92\linewidth]{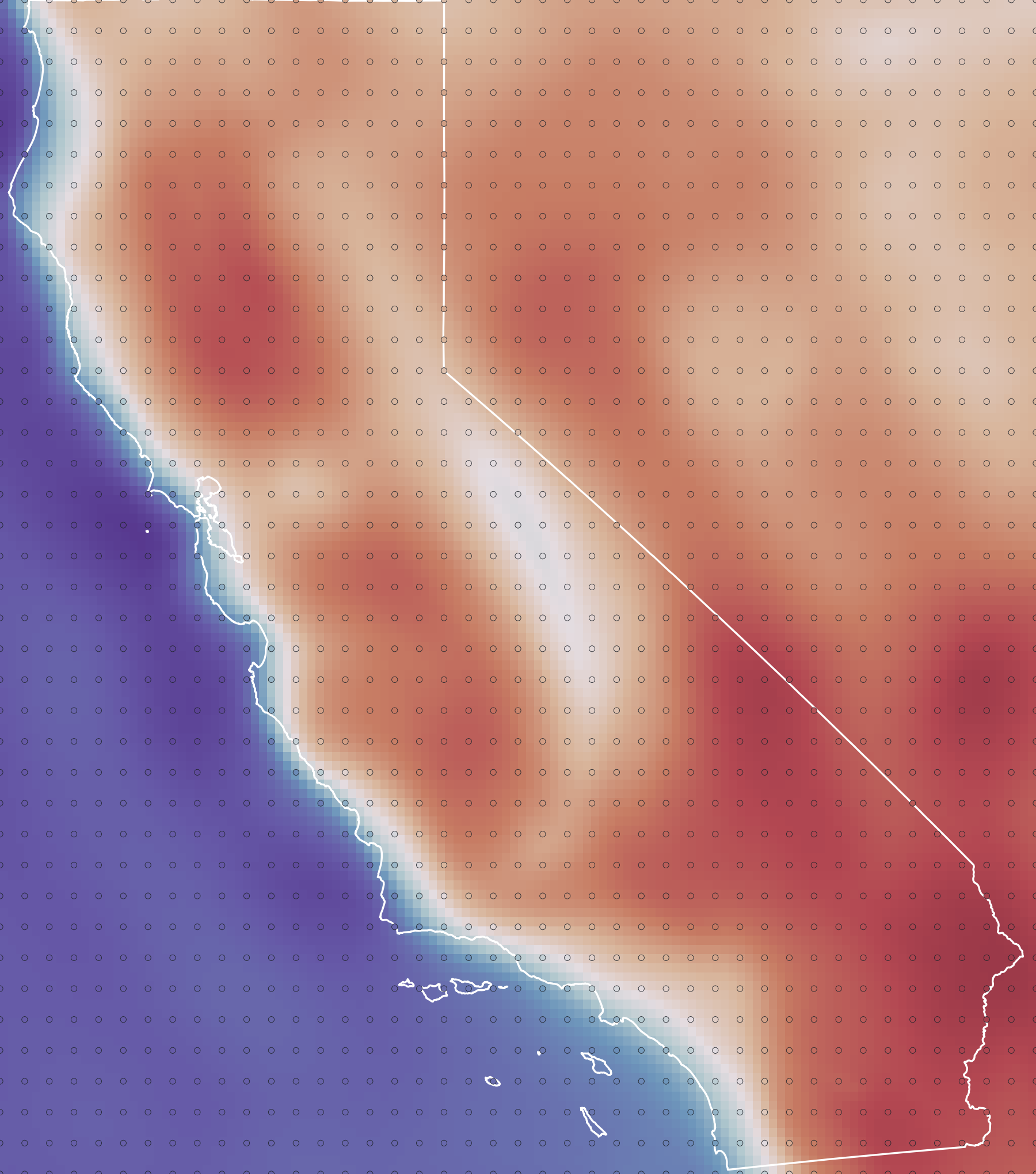}
    \makeatletter\def\@captype{figure}\makeatother
    \caption{\small\textbf{Temperature heat map of California} Circles show the SFNO's temperature predictions at ERA5's grid resolution of $0.25^\circ$. Our framework extends beyond the grid, providing maximum temperature bounds over the entire region of California.}
    \label{fig:sfno-temperature}
\end{minipage}\hfill
\begin{minipage}[t]{0.54\textwidth}
\vspace{0pt}
Finally, we demonstrate that our certification framework scales to foundation-scale weather forecasting models. Neural operator-based models have achieved competitive accuracy and substantial speedups in global weather forecasting~\citep{kurth2022fourcastnetacceleratingglobalhighresolution,bonev2025fourcastnet3geometricapproach,bonev2023sphericalfourierneuraloperators}. In particular, the Spherical Fourier Neural Operator (SFNO)~\citep{bonev2023sphericalfourierneuraloperators} replaces the FNO's FFT with spherical harmonic transforms to model global atmospheric dynamics. We evaluate the pretrained SFNO from NVIDIA Earth2Studio~\citep{Earth2Studio_Contributors_NVIDIA_Earth2Studio_2024}, which is trained on the Earth Reanalysis
5 (ERA5) dataset~\citep{hersbach2020era5} to predict 6-hour global weather evolution across 73 atmospheric variables at $0.25^\circ$ resolution ($721\times 1440$ latitude--longitude grid).

\vspace{5pt}
Certifying this model is challenging due to its sheer scale. The pretrained SFNO contains 572.5 million parameters across 8 layers and 384 hidden channels. Each layer couples global spherical harmonic transforms up to degree 239 with dense channel mixing and normalization over more than one million spatial grid points per channel. For this model, we focus on verifying the maximum output over a region. Specifically, the maximum predicted 2-meter temperature over California:
\[
\max_{(\varphi,\lambda)\in R}T_\theta(\varphi,\lambda),
\quad
R=[32.5^\circ,42^\circ]\times[-124.5^\circ,-114^\circ],
\]
\end{minipage}
\par
\vspace{5pt} where \(\varphi\) and \(\lambda\) denote latitude and longitude, and \(T_\theta\) is the six-hour temperature forecast in degrees Celsius at 5:00 p.m. PDT on July 15, 2023. Using function-space IBP with Clenshaw–Curtis quadrature, representing the fixed ERA5 input by a spherical harmonic expansion up to degree 239, we obtain
\[
\max_{(\varphi,\lambda)\in R}T_\theta(\varphi,\lambda)
\;\le\;47.240^\circ\mathrm C.
\]
The certified interval is only \(0.0396^\circ\mathrm C\) wide, showing that the bound is tight, and covers every location in the continuous region, including points between grid locations. The full calculation takes 3.64 hours on a single NVIDIA B300 GPU. For comparison, the sampled forecast maximum of the SFNO is \(47.194^\circ\mathrm C\). Figure~\ref{fig:sfno-temperature} shows the sampled temperature predictions and the certified upper bounds across the region.
\section{Discussion and Limitations}
In this work, we certify each PDE in its \emph{strong form} by bounding the equation residual at every point across the continuous domain. This contrasts with weak-form methods, which check weighted averages of the equation against smooth test functions so that fewer derivatives are needed~\citep{kharazmi2021hpvpinns,zang2020weak}. Our post-training approach is also distinct from classical numerical analysis: rather than proving ahead of time that a numerical scheme converges as the grid is refined~\citep{courant1928partiellen,lax1956survey}, we directly certify a trained neural operator on a specific input or input set.

Because function-space certification provides guarantees over the entire continuous domain, it requires an input function that is defined everywhere, together with bounds on its spatial derivatives. This requirement is natural when the forcing term or initial condition is given by a mathematical formula, but it poses a limitation when only discrete measurements are available, as in the gridded ERA5 weather dataset. In such cases, one must first construct a continuous interpolant (such as our spherical harmonic expansion for ERA5) before running certification, and the resulting certificate applies to that chosen interpolant rather than to unknown values between measurement points.

Finally, each certified property faces practical scaling limits. Like standard bound propagation in finite dimensions, the looseness of our bounds grows exponentially with network depth when propagating input perturbations. This depth growth is why we do not certify adversarial robustness for the 8-layer SFNO or across multi-step rollouts of the KS equation, and instead verify fixed-input properties such as maximum output and PDE residuals, where errors compound more slowly across layers. Similarly, residual certification is only possible when a known closed-form PDE governs the system, which excludes purely data-driven tasks such as global weather forecasting. Even when a governing PDE is known, converting a tight residual bound into a solution-error bound requires an analytical stability estimate. For chaotic systems such as the KS equation, worst-case stability bounds grow exponentially over time (Appendix~\ref{app:ks-stability}) and are too conservative to yield informative solution guarantees.

\section{Conclusion}
We propose the mathematical framework for bound propagation
through function-to-function layers, and its implementation for the neural operator. By combining bound functions with rigorous numerical integration, our approach certifies model behavior over the continuous domain rather than only at a fixed set of grid points. Experimental results over a broad range of models and verification problems show that our framework is both efficient (under one hour on a single GPU for most certification tasks) and tight. 

A promising direction is extending function-space robustness certification to more complex algorithms, such as $\alpha$-CROWN~\citep{xu2021fast}, $\beta$-CROWN~\citep{wang2021beta}, and branch-and-bound methods~\citep{shi2025genbab} to obtain tighter bounds, as well as certifying other physical properties, such as safety, physical constraints, and conservation laws. Beyond mathematical soundness, accounting for floating-point errors in tensor operations by scaling Arb for full-scale certification can provide guarantees to machine-level precision. Together, these contributions bridge the gap between the speed of neural surrogates and the worst-case guarantees of classical solvers, paving the way for the deployment of neural operators in safety-critical scientific and engineering applications.

\subsection*{Data availability}
Datasets and models used in this study are available at \url{https://github.com/neuraloperator/function-space-certification-no}
\subsection*{Code availability}
Source code is available at \url{https://github.com/neuraloperator/function-space-certification-no}

\subsection*{Acknowledgments}
Ryan Hsiang is supported by the Summer Undergraduate Research Fellowships (SURF) program at Caltech. Robert Joseph George is supported by DARPA expMath and AA2.AIMING-1-NSF.2522494. Anima Anandkumar is supported in part by the Bren endowed chair, DARPA expMath, and the AI2050 Senior Fellowship program at Schmidt Sciences.

\subsection*{AI use statement}
Large language models were used to debug code, write unit tests, and polish the writing and grammar of the manuscript. The core algorithms and main experiments were designed and implemented by the authors, and the correctness of the results was verified through detailed testing and manual review. The authors take full responsibility for the manuscript and the codebase.

\bibliographystyle{unsrtnat}
\bibliography{main}

\clearpage
\appendix
\startcontents[appendix]
\section*{Appendix contents}
\printcontents[appendix]{}{1}{\setcounter{tocdepth}{2}}

\section{Related Work}

\paragraph{Neural network verification.}
Neural network verification checks whether a model's output stays within a safe range for every input in a chosen region, rather than only for a finite set of test points. Early methods formulated verification as a logical constraint or mixed-integer optimization problem, using satisfiability modulo theories (SMT) solvers such as Reluplex~\citep{katz2017reluplexefficientsmtsolver} to check piecewise-linear networks with rectified linear unit (ReLU) activations. While these solvers can give exact yes-or-no answers, checking every possible ReLU on/off combination becomes too slow as networks grow larger.

To scale to larger models, \emph{bound propagation} methods compute guaranteed lower and upper bounds layer by layer~\citep{gowal2019effectivenessintervalboundpropagation,singh2018fast,zhang2018efficient}. Interval bound propagation (IBP)~\citep{gowal2019effectivenessintervalboundpropagation} tracks a simple min-to-max range for each neuron, whereas linear relaxation methods such as CROWN~\citep{zhang2018efficient} replace nonlinear activations with linear lower and upper bounds to keep track of how neurons depend on one another. Auto-LiRPA~\citep{xu2020automatic} generalized this linear bound propagation to general computational graphs on GPUs, and optimizing the slopes of the linear bounds ($\alpha$-CROWN) made fast, GPU-parallel verification practical~\citep{xu2021fast}. Because replacing nonlinear layers with linear bounds introduces some looseness~\citep{salman2019convex}, \emph{complete} verifiers combine fast bound propagation with branch-and-bound (BaB) search, which splits uncertain neurons into separate cases to refine the bounds. In particular, $\beta$-CROWN~\citep{wang2021beta} encodes ReLU split constraints directly into the bound-propagation pass with learnable multipliers, and GCP-CROWN~\citep{zhang2022general} adds general cutting-plane constraints that capture relationships across multiple neurons at once; recent extensions further speed up branching and tighten constraints automatically~\citep{zhou2024scalable,zhou2025clip,davis2026lookaheadbranchingneuralnetwork}.

Beyond standard ReLU classification, bound propagation and branch-and-bound have been extended to general smooth nonlinearities and architectures~\citep{shi2025genbab}, to bounding the set of inputs (the preimage) that map to a target output region~\citep{kotha2023provably}, and to bounding network derivatives---such as local Lipschitz constants (bounds on how fast the output can change with respect to the input)~\citep{shi2022efficiently} and higher-order partial derivatives in physics-informed neural networks ($\partial$-CROWN)~\citep{eiras2024efficienterrorcertificationphysicsinformed}. Separately, interactive theorem provers such as Lean have been connected to neural network verification through frameworks like TorchLean~\citep{george2026torchleanformalizingneuralnetworks}, which checks network computations and verification proofs down to foundational mathematical axioms. However, all of these tools operate on finite-dimensional vectors, whereas our work extends IBP and CROWN to infinite-dimensional function spaces.
\paragraph{Robustness of neural operators.}
Neural operators such as FNOs perform well on parametric PDE benchmarks, but their behavior under adversarial or distributional perturbation is comparatively understudied. A systematic study of FNO adversarial robustness~\citep{adesoji2022evaluatingadversarialrobustnessfourier} shows that accuracy degrades rapidly under norm-bounded perturbations on Darcy and Navier--Stokes equations. Approximation and sample-complexity guarantees for FNOs applied to dissipative evolution operators admitting stable spectral discretizations are established by~\citep{chandramoorthy2026spectralmethodssamplecomplexity}. These works expose sources of fragility in neural operators but stop short of providing deterministic, per-instance guarantees.

\citet{ma2024calibrateduncertaintyquantificationoperator} train a neural operator to predict uncertainty bands and use conformal prediction to calibrate them using examples with known ground truth. Its guarantee is:
\[
\Pr_{(a,u)}
\left[
\Pr_x\!\left(
\|u(x)-{\mathcal G_\theta}(a)(x)\|_2
\le\lambda E(a)(x)
\right)
<1-\alpha
\right]
\le\delta.
\]
This provides an uncertainty estimate similar to our PDE solution certification, but their guarantee is probabilistic, compared to the worst-case guarantee we provide.

\paragraph{Certified and formally verified PDE surrogates.}
A separate line of work seeks to certify the error of a trained surrogate. Other than $\partial$-CROWN~\citep{eiras2024efficienterrorcertificationphysicsinformed}, wavelet-based PINNs with a computable a posteriori error bound derived from a weighted adaptive wavelet expansion of the residual are constructed by~\citep{ernst2025certified}. Certification at the level of function-space norms, combining interval-arithmetic enclosures with adaptive quadrature to obtain certified $L^p$ and Sobolev-norm bounds, with an explicit application to PINN residuals, is studied by~\citep{grundler2026certifiedfunctionspace}. Naive, bound-propagation-based (IBP/CROWN-style), and interval-weight architectures for propagation of interval bounds through MLP and DeepONet surrogates are compared by~\citep{faza2026directinterval}. In Lean, TorchLean~\citep{george2026torchleanformalizingneuralnetworks} provides the training and verification workflows for FNOs directly in Lean. FNOs can also be compiled into an SMT representation as a piecewise-linear map, checking properties such as positivity and mass conservation~\citep{baheri2026formallyverifyneuralpde}. However, these methods are limited to fixed-grid verification, similar to our grid verification methods. Relative to this line of work, we target certified robustness verification for FNOs at scales beyond what current SMT- or interval-arithmetic-based approaches support.

\paragraph{System-level and agentic use of neural operators.}
Beyond single-shot prediction, local neural operators have been embedded within Equation-Free, Krylov-subspace methods to perform system-level fixed-point, stability, and bifurcation analysis directly from data~\citep{fabiani2026enabling}. As LLM agents are increasingly used to automate SciML pipelines, error-metric-based model selection has been argued to be insufficient, motivating a verification-first workflow that checks every candidate surrogate against machine-checkable physics requirements before it is reported as viable~\citep{abueidda2026physicsauditedagenticdiscoveryscientific}.

\paragraph{Classical numerical guarantees.}
Classical numerical analysis provides a well-developed framework for reasoning
about the behavior of numerical approximations under mesh refinement, including
notions of consistency, stability, and convergence
~\citep{lax1956survey,leveque2007finite,quarteroni1994numerical}.
For some time-dependent discretizations, stability additionally requires
scheme-specific conditions such as the CFL condition
~\citep{courant1928partiellen}. Spectral methods admit analogous approximation
guarantees, with convergence rates determined by the regularity and Fourier
coefficient decay of the represented functions
~\citep{canuto2006spectral,trefethen2000spectral}.

Neural operators differ from a conventional discretization in that they are
formulated as mappings between function spaces and can share the same learned
parameters across multiple spatial resolutions
~\citep{kovachki2023neuraloperator}. Nevertheless, an FNO must still be
evaluated numerically on a finite grid. Recent analysis quantifies the
resulting discretization error and establishes convergence of finite-grid FNO
evaluations toward their continuum counterparts under suitable regularity
assumptions~\citep{lanthaler2025discretizationerrorfourierneural}.

Our certification problem is related but distinct. Let $N$ denote the
resolution at which a discrete FNO is evaluated. Changing $N$ changes the
finite-dimensional approximation used to represent the continuum operator.
In contrast, our verifier introduces a separate quadrature resolution $Q$.
The quantity $Q$ controls the numerical accuracy with which we compute a
certificate for the \emph{same} continuum FNO; it does not define the operator
being certified. The quadrature approximation error is explicitly enclosed by our integration remainders (Appendix~\ref{app:crown-quadrature}), which decrease as $\mathcal O(Q^{-1})$ for non-smooth integrands and geometrically with order for Gauss--Legendre and Clenshaw--Curtis quadrature.

Thus the two refinement procedures play different roles:
\[
\begin{array}{lll}
N\uparrow
&:&
\text{refines the finite-grid representation of the FNO},\\[2mm]
Q\uparrow
&:&
\text{tightens a sound numerical enclosure of the same continuum FNO}.
\end{array}
\]
Function-space certification therefore complements discretization-convergence
analysis: rather than certifying a single sampled computation, it constructs
bounds that remain valid for the continuum operator over the full spatial
domain.

\section{Certified integration methods}
\label{app:crown-quadrature}

Every spectral layer of a neural operator computes an integral of an intermediate function against a learned kernel. A computer cannot evaluate a continuous integral directly; instead, numerical integration (often called \emph{quadrature}) approximates the integral by a weighted sum of function values at $n$ sample points $x_1,\ldots,x_n$~\citep{NIST:DLMF}:
\begin{equation}
    \int_a^b f(x)\,dx = \sum_{k=1}^n w_k f(x_k) + E_n(f).
    \label{eq:interpolatory-quadrature}
\end{equation}
Here $x_k$ are the sample points (nodes), $w_k$ are the weights, and $E_n(f)$ is the approximation error (remainder). Standard quadrature rules choose the weights $w_k$ by fitting a polynomial of degree $n-1$ through the $n$ sample points and integrating that polynomial exactly; consequently, the error $E_n(f)$ is zero whenever $f$ is a polynomial of degree at most $n-1$.

In standard scientific computing, the remainder $E_n(f)$ is simply dropped. For formal certification, however, dropping the remainder is unsound: we must bound both the function values $f(x_k)$ and the worst-case magnitude of the remainder $|E_n(f)|$. Below we give the non-smooth and smooth error bounds for Gauss--Legendre quadrature and the Clenshaw--Curtis rule used on the sphere.

\subsection{Gauss--Legendre quadrature}
\label{app:gauss-legendre}

Numerical integration (or \emph{quadrature}) approximates a definite integral by a weighted average of function evaluations at $n$ sample points:
\[
    \int_{-1}^1 f(x)\,dx \approx \sum_{j=1}^n w_j f(x_j).
\]
In basic calculus rules, where nodes are uniformly placed, the weights are
\[
x_j=-1+(j-1)h,\qquad h=\frac{2}{n-1},\qquad
w_j=
\begin{cases}
    h/2, & j=1 \text{ or } j=n,\\
    h,   & 2\le j\le n-1.
\end{cases}
\]
\emph{Gauss--Legendre quadrature}~\citep{NIST:DLMF} optimizes both the sample locations $x_j$ and the weights $w_j$, such that $n$ sample points integrate all polynomials of degree up to $2n-1$ with zero error. The nodes and weights are defined by
\[
P_n(x_j)=0,\quad w_j=\frac{2}{(1-x_j^2)[P_n'(x_j)]^2}.
\]
where $P_n(x)$ are the \emph{Legendre polynomials}, given by,
\[
    P_n(x) = \frac{1}{2^n n!}\frac{d^n}{dx^n}(x^2-1)^n.
\]
A consequence of this is that the weights sum to $2$,
\begin{align*}
    \sum_{j=1}^n w_j=\int_{-1}^{1}1\,dx=2.
\end{align*}

For an arbitrary function $f$ over $[a,b]$, Cauchy's estimate says that if $f$ is analytic and $|f(z)|\leq M$ over a complex disk $|z-c|\leq R$, then it has a convergent Taylor expansion with
\begin{align*}
    f(z) = \sum_{k=0}^\infty a_k(z-c)^k,\quad |a_k|\leq \frac{M}{R^k}.
\end{align*}
where $c=\frac{a+b}{2}$, and $R>h=(b-a)/2$. Let $q_m$ be the Taylor polynomial through degree $m$, then
\[
\begin{aligned}
|f(x)-q_{2n-1}(x)|
&\le M\sum_{k=2n}^\infty\left(\frac hR\right)^k=\frac{M(h/R)^{2n}}{1-h/R}.
\end{aligned}
\]
Let $r(x)=f(x)-q_{2n-1}(x)$, since Gaussian quadrature integrates $q_{2n-1}$ exactly, the integration error is,
\begin{equation}
    \left|\int_{a}^br(x)\,dx- \sum_{j=1}^n w_j r(x_j)\right|\leq4h\|r\|_\infty\leq\frac{4hM(h/R)^{2n}}{1-h/R}.
    \label{eq:gauss-derivative-remainder}
\end{equation}
Choosing $R=4h$, the integration error decays exponentially,
\begin{align*}
    \frac{4hM(h/R)^{2n}}{1-h/R}=\frac{16}{3}hM4^{-2n}.
\end{align*}
In our implementation, we subdivide integration panels until analyticity and a finite magnitude bound can be certified on the corresponding complex disks of radius \(4h\).
\subsection{Clenshaw--Curtis quadrature}
\label{app:clenshaw-curtis}

While Gauss--Legendre quadrature achieves the highest possible accuracy for a fixed number of points, its sample points are unevenly spaced, have no closed-form formula, and exclude the endpoints $\pm 1$. \emph{Clenshaw--Curtis quadrature}~\citep{Clenshaw:1960:MNI,doi:10.1137/060659831} is an alternative high-order method on $[-1,1]$ that instead samples the function at $n+1$ equally spaced angles $\theta_k$ (including both endpoints):
\begin{equation}
    x_k = \cos\theta_k,
    \qquad
    \theta_k = \frac{k\pi}{n},
    \qquad k=0,1,\ldots,n.
    \label{eq:cc-nodes}
\end{equation}

The method is based on the change of variables $x=\cos\theta$ for $\theta\in[0,\pi]$. Since $dx = -\sin\theta\,d\theta$, the integral on $[-1,1]$ becomes an integral over $[0,\pi]$:
\begin{equation}
    \int_{-1}^1 f(x)\,dx
    = \int_0^\pi f(\cos\theta)\sin\theta\,d\theta.
    \label{eq:cc-theta-transform}
\end{equation}
This substitution makes Chebyshev polynomials $T_m(x)=\cos(m\cos^{-1} x)$ easy to work with, as they turn into simple cosine waves $T_m(\cos\theta)=\cos(m\theta)$. Thus, expanding $f(x)$ in Chebyshev polynomials is the same as writing $f(\cos\theta)$ as a Fourier cosine series:
\[
    f(\cos\theta) = \frac{a_0}{2} + \sum_{m=1}^\infty a_m\cos(m\theta).
\]
To integrate $f(x)$, we only need to integrate each cosine term against $\sin\theta$, which has a simple closed form:
\begin{equation}
    \int_{-1}^1 T_m(x)\,dx
    = \int_0^\pi \cos(m\theta)\sin\theta\,d\theta
    = \begin{cases}
        \dfrac{2}{1-m^2}, & m\text{ even},\\[6pt]
        0, & m\text{ odd}.
    \end{cases}
    \label{eq:chebyshev-moments}
\end{equation}

In practice, we evaluate $f$ at the $n+1$ equally spaced angles $\theta_k$ in \eqref{eq:cc-nodes}, fit a cosine sum through the sample values $f(x_k)$, and integrate term-by-term using \eqref{eq:chebyshev-moments}. Grouping the terms by each sample value $f(x_k)$ gives an explicit formula for the quadrature weights $w_k$~\citep{NIST:DLMF}:
\begin{equation}
    \int_{-1}^1 f(x)\,dx \approx \sum_{k=0}^n w_k f(x_k),
    \qquad
    w_k = \frac{g_k}{n}\left(
        1 - \sum_{j=1}^{\lfloor n/2\rfloor}
        \frac{b_j}{4j^2-1}\cos\!\left(\frac{2jk\pi}{n}\right)
    \right),
    \label{eq:cc-weights}
\end{equation}
where $g_k=1$ at the endpoints ($k=0,n$) and $g_k=2$ at interior points ($0 < k < n$), and $b_j=1$ if $j=n/2$ and $b_j=2$ otherwise. All weights $w_k$ are positive, sum to $2$, and integrate every polynomial of degree up to $n$ exactly~\citep{doi:10.1137/060659831}.

Clenshaw--Curtis quadrature fits naturally with spherical models like the Spherical Fourier Neural Operator (SFNO) because its sample angles match standard latitude--longitude grids. On the unit sphere, each point is given by colatitude $\theta\in[0,\pi]$ (the angle from the North Pole) and longitude $\phi\in[0,2\pi)$, with surface area element $d\Omega = \sin\theta\,d\theta\,d\phi$.

For colatitude $\theta$, the spherical integral $\int_0^\pi (\cdots)\sin\theta\,d\theta$ has the same $\sin\theta$ weight as \eqref{eq:cc-theta-transform}. We use $Q_\theta=n+1$ evenly spaced points $\theta_k=k\pi/n$ from the North Pole to the South Pole, consistent with the latitude grids used in weather datasets such as ERA5. For longitude $\phi$, the function is $2\pi$-periodic, so we apply the trapezoidal rule with $Q_\phi$ evenly spaced points $\phi_j=2\pi j/Q_\phi$, each with weight $2\pi/Q_\phi$~\citep{NIST:DLMF}.

Combining Clenshaw--Curtis in latitude with the trapezoidal rule in longitude lets us compute each spherical harmonic coefficient $\widehat z_{\ell m}$ of a hidden field $z(\theta,\phi)$ directly on the uniform latitude--longitude grid:
\begin{equation}
    \widehat z_{\ell m}
    = \int_0^{2\pi}\!\!\int_0^\pi z(\theta,\phi)\,\overline{Y_\ell^m(\theta,\phi)}\,\sin\theta\,d\theta\,d\phi
    \approx \frac{2\pi}{Q_\phi}\sum_{k=0}^n w_k P_\ell^m(\cos\theta_k)
    \sum_{j=0}^{Q_\phi-1} z(\theta_k,\phi_j)e^{-\mathrm{i}m\phi_j}.
    \label{eq:sfno-cc-quadrature}
\end{equation}
To bound the approximation error of \eqref{eq:sfno-cc-quadrature}, we take advantage of the fact that it samples at evenly spaced angles. First, consider a $2\pi$-periodic function $f(x+2\pi)=f(x)$ with integral $I = \int_0^{2\pi} f(x)\,dx$, approximated by the $Q$-point periodic trapezoidal rule:
\[
    I_Q = \frac{2\pi}{Q}\sum_{j=0}^{Q-1} f\!\left(\frac{2\pi j}{Q}\right).
\]
Expanding $f$ in a Fourier series $f(x) = \sum_{k\in\mathbb{Z}} \widehat f_k e^{\mathrm{i}kx}$ with coefficients $\widehat f_k = \frac{1}{2\pi}\int_0^{2\pi} f(x)e^{-\mathrm{i}kx}\,dx$. We can show that,
\begin{equation}
I = 2\pi\widehat f_0,\quad
    I_Q = 2\pi\sum_{m\in\mathbb{Z}} \widehat f_{mQ}
    \quad\Longrightarrow\quad
    I_Q - I = 2\pi\sum_{m\ne 0} \widehat f_{mQ}.
    \label{eq:trapezoidal-aliasing}
\end{equation}
Similar to the Gaussian quadrature error bounds, we assume $f$ is analytic to a complex strip $|\operatorname{Im}z| < \rho$ with $|f(z)|<M$, then
\[
    \widehat f_k = \frac{1}{2\pi}\int_0^{2\pi} f(x-\mathrm{i}\rho)e^{-\mathrm{i}k(x-\mathrm{i}\rho)}\,dx.
\]
and
\[
    |\widehat f_k|
    \le \frac{1}{2\pi}\int_0^{2\pi} |f(x-\mathrm{i}\rho)|e^{-k\rho}\,dx
    \le M e^{-k\rho}.
\]
It follows that
\begin{equation}
    |I_Q - I|
    \le 2\pi M\sum_{m\ne 0} e^{-\rho|mQ|}
    = 4\pi M\sum_{m=1}^\infty e^{-\rho mQ}
    = \frac{4\pi M}{e^{\rho Q} - 1}
    \label{eq:strip-error-bound}
\end{equation}
Thus, the certified error again decreases exponentially with the grid resolution $Q$.

For latitude, the error follows from the cosine expansion used to construct Clenshaw--Curtis quadrature. Write the longitude-integrated function as
\[
    g(\theta)=\int_0^{2\pi}
    z(\theta,\phi)\,\overline{Y_\ell^m(\theta,\phi)}\,d\phi
    =\frac{a_0}{2}+\sum_{k=1}^{\infty}a_k\cos(k\theta).
\]
The spherical coordinate identity $(\theta,\phi)\equiv(-\theta,\phi+\pi)$ makes $g$ even, and its extension in $\theta$ is $2\pi$-periodic. The coefficient to be computed is
\[
    \widehat z_{\ell m}=\int_0^\pi g(\theta)\sin\theta\,d\theta.
\]
Assume that $g$ extends analytically to $|\operatorname{Im}\theta|<\rho_\theta$, with $|g|\le M_\theta$ throughout this strip. Its cosine coefficients then satisfy
\[
    |a_k|\le 2M_\theta e^{-\rho_\theta k}.
\]

Let $n=Q_\theta-1$. At the quadrature nodes $\theta_j=j\pi/n$, each higher-frequency cosine agrees with a lower-frequency cosine:
\[
    \cos(k\theta_j)=\cos(r_n(k)\theta_j),
    \qquad
    r_n(k)=\min\{k\bmod 2n,\;2n-(k\bmod 2n)\}.
\]
Let $J_k$ denote the cosine integral in \eqref{eq:chebyshev-moments}. Clenshaw--Curtis assigns the $k$-th cosine term the integral $J_{r_n(k)}$ instead of $J_k$. All terms with $k\le n$ are integrated exactly, so the latitude error satisfies
\begin{equation}
    E_\theta
    =\left|\sum_{k=n+1}^{\infty}a_k\bigl(J_k-J_{r_n(k)}\bigr)\right|
    \le
    2M_\theta\sum_{k=n+1}^{\infty}
    e^{-\rho_\theta k}\left|J_k-J_{r_n(k)}\right|.
    \label{eq:cc-latitude-error}
\end{equation}
Odd $k$ contribute zero, so only even frequencies need to be included.

Finally, replacing the exact longitude integral by its periodic trapezoidal approximation introduces a second error. Assume that $z\overline{Y_\ell^m}$ is analytic in a longitude strip of width $\rho_\phi$, with magnitude at most $M_\phi$ uniformly over real latitudes. Equation~\eqref{eq:strip-error-bound} and the positive latitude weights, which sum to $2$, give
\begin{equation}
    E_\phi\le\frac{8\pi M_\phi}{e^{\rho_\phi Q_\phi}-1}.
    \label{eq:cc-longitude-error}
\end{equation}
By the triangle inequality, the total coefficient error satisfies
\begin{equation}
    \left|\widehat z_{\ell m}-\widehat z_{\ell m}^{\,Q}\right|
    \le E_\theta+E_\phi,
    \label{eq:cc-total-error}
\end{equation}
where $\widehat z_{\ell m}^{\,Q}$ is the quadrature approximation in \eqref{eq:sfno-cc-quadrature}.

\subsection{Non-smooth error bound}
\label{app:quad-constants}

The smooth Cauchy bound relies on extending the integrand into the complex plane. When a neural operator uses a non-smooth activation such as ReLU, or when the input function can vary arbitrarily within an uncertainty band, the integrand can have kinks and is no longer analytic. In this case, high-order derivative bounds are unavailable, so we bound the integration error using only the maximum slope (first derivative) of the integrand on each cell.

As in Section~\ref{sec:certified-integration}, we partition the domain $\Omega=[0,L)$ into $Q$ uniform cells $C_j=[x_j,x_j+h)$ of width $h=L/Q$. On each cell $C_j$, we apply the two-point Gauss--Legendre rule. Rescaling the roots $\pm 1/\sqrt{3}$ of $P_2(x)=\frac{1}{2}(3x^2-1)$ from $[-1,1]$ to $C_j$ places the two sample points at
\begin{equation}
    x_{j,k} = x_j + t_k h,
    \qquad
    t_1 = \frac{1}{2} - \frac{1}{2\sqrt{3}},
    \qquad
    t_2 = \frac{1}{2} + \frac{1}{2\sqrt{3}},
    \label{eq:gauss-nodes}
\end{equation}
each with weight $h/2$. Suppose $f:\Omega\to\mathbb{R}$ is continuous on $C_j$ with maximum slope $\sup_{s\in C_j}|f'(s)|\le M_{f,j}$. We split the cell at the midpoint $x_j+h/2$: on the left half $[x_j,x_j+h/2)$ we compare $f(s)$ to the first sample value $f(x_{j,1})$, and on the right half $[x_j+h/2,x_j+h)$ we compare it to $f(x_{j,2})$. By the mean value theorem, $|f(s)-f(x_{j,k})|\le M_{f,j}|s-x_{j,k}|$ on each half-cell. Integrating this distance over the two half-cells gives the single-cell error bound
\begin{equation}
    \left|\int_{C_j} f(s)\,ds - \frac{h}{2}\bigl(f(x_{j,1})+f(x_{j,2})\bigr)\right|
    \le M_{f,j}\left(
        \int_{x_j}^{x_j+h/2} |s-x_{j,1}|\,ds
        + \int_{x_j+h/2}^{x_j+h} |s-x_{j,2}|\,ds
    \right)
    = d\,h^2 M_{f,j},
    \label{eq:cell-gauss-error}
\end{equation}
where the transport constant is
\[
    d
    = t_1^2 + \left(\frac{1}{2}-t_1\right)^2
    = \frac{5}{12} - \frac{\sqrt{3}}{6}
    \approx 0.128.
\]
Geometrically, $d\,h$ is the average distance from a point in $C_j$ to its nearest Gauss node (compared to $d=1/2$ for the left-endpoint rectangle rule). Summing \eqref{eq:cell-gauss-error} over all $Q$ cells and replacing each exact node value $f(x_{j,k})$ with an interval enclosure $[\underline f_{j,k},\overline f_{j,k}]$ yields the certified quadrature rule \eqref{eq:corrected-quadrature-proposition}.

In our propagation rules, the hidden field $z$ is stored at the cell left endpoints $x_j$ together with a cellwise slope bound $M_{z,j}=\max\{|\underline z'_j|,|\overline z'_j|\}$ satisfying $|z'(s)|\le M_{z,j}$ on $C_j$. Rather than storing extra values at the interior Gauss points $x_{j,k}$, we enclose $z(x_{j,k})$ directly from the left endpoint using the slope bound:
\begin{equation}
    \underline z_{j,k} = \underline z(x_j) - t_k h M_{z,j},
    \qquad
    \overline z_{j,k} = \overline z(x_j) + t_k h M_{z,j}.
    \label{eq:gauss-node-enclosure}
\end{equation}
Similarly, over the entire cell $C_j$, $z(s)$ lies in $[\underline z(x_j)-hM_{z,j},\,\overline z(x_j)+hM_{z,j}]$, so its magnitude on $C_j$ is bounded by
\[
    S_{z,j} = \max\bigl\{|\underline z(x_j)-hM_{z,j}|,\,|\overline z(x_j)+hM_{z,j}|\bigr\}.
\]
For a spectral convolution integrand $f(s)=\kappa(x_q-s)z(s)$ at target node $x_q$, the product rule gives $|f'(s)|\le |\kappa'(x_q-s)|\,|z(s)| + |\kappa(x_q-s)|\,|z'(s)|$. If $S_{\kappa,qj}$, $S_{\kappa',qj}$, and $S_{\kappa'',qj}$ bound $|\kappa|$, $|\kappa'|$, and $|\kappa''|$ for $s\in C_j$, the cellwise slope of the integrand is bounded by
\[
    M_{f,j} = S_{\kappa',qj} S_{z,j} + S_{\kappa,qj} M_{z,j}.
\]
Substituting $M_{f,j}$ into \eqref{eq:corrected-quadrature-proposition} gives the forward quadrature remainder $\eta$, and replacing $\kappa$ with $\kappa'$ gives the derivative remainder $\eta'$:
\begin{equation}
    \eta = d\,h^2\sum_{j=0}^{Q-1}\bigl(S_{\kappa',qj}S_{z,j} + S_{\kappa,qj}M_{z,j}\bigr),
    \qquad
    \eta' = d\,h^2\sum_{j=0}^{Q-1}\bigl(S_{\kappa'',qj}S_{z,j} + S_{\kappa',qj}M_{z,j}\bigr).
    \label{eq:explicit-forward-constant}
\end{equation}
The same product-rule bound is used in the backward CROWN pass \eqref{eq:crown-spectral-derivative-bound}, with the coefficient field $c$ in place of $z$. For multi-dimensional domains, the rule is applied as a tensor product across coordinate axes, and the remainder sums the partial-derivative contributions along each axis.

\subsection{Ablation study of integration methods}
\label{app:integration-ablation}

To motivate our choice of Gauss--Legendre quadrature, we compare it against two baselines: the left rectangle rule, which is the simplest rule with a slope-based error bound, and the Arb library~\citep{johansson2018numericalintegrationarbitraryprecisionball}, which computes rigorous integrals in arbitrary-precision ball arithmetic. A key feature of Arb is that it is sound down to floating-point error, whereas our quadrature rules are not. However, its practicality is limited by its CPU-based implementation. We evaluate all three on three FNOs with zero input uncertainty ($\varepsilon=0$): 1D Poisson with 1 layer and 4 channels, 1D Poisson with 2 layers, 32 channels, and 32 modes, and 2D Poisson with 2 layers and 8 channels. We use tanh activation for all models. We measure the maximum pointwise output-bound width across 64 fixed query points ($x_j=j/64$ in 1D and $(i/8,j/8)$ in 2D). Table~\ref{tab:integration-ablation} reports the results.

 We allocate a compute budget of 10 and 60 seconds for the benchmarks. However, Arb did not finish the 2D Poisson benchmark within the allocated time and is estimated to take 5-6 days, showing its limited scalability.
Gaussian quadrature achieves bounds many orders of magnitude tighter than rectangle quadrature at similar runtime, and matches or improves on Arb while running on the GPU. Under input perturbation ($\varepsilon=10^{-3}$), where both rules use the slope-based error bound on the ReLU models at $Q=2^{20}$, the integration error is a small fraction of the bound width, and Gaussian quadrature gives the same or slightly tighter bounds than rectangle quadrature (by up to $1.7\%$) at similar cost.

\begin{table}[!htbp]
\centering
\caption{Comparison of certified integration backends across 64 query points ($\varepsilon=0$).}
\label{tab:integration-ablation}
\begin{tabular}{llrrl}
\toprule
Benchmark & Method & Max width $\downarrow$ & Runtime & Device\\
\midrule
1D (1 layer)
& Rectangle & $6.27\times 10^{-6}$ & $4.92$s & GPU\\
& Gaussian & $8.56\times 10^{-76}$ & $9.58$s & GPU\\
& Arb & $2.46\times 10^{-71}$ & $9.16$s & CPU\\
\midrule
1D (2 layers)
& Rectangle & $3.29\times 10^{-2}$ & $49.23$s & GPU\\
& Gaussian & $1.72\times 10^{-187}$ & $53.34$s & GPU\\
& Arb & $2.39\times 10^{-14}$ & $53.90$s & CPU\\
\midrule
2D (2 layers)
& Rectangle & $3.02$ & $46.81$s & GPU\\
& Gaussian & $1.29\times 10^{-106}$ & $25.13$s & GPU\\
& Arb & Did not finish & 5--6 days & CPU\\
\bottomrule
\end{tabular}
\end{table}

\section{Function-space IBP propagation}
\label{app:ibp-rules}

This section gives the rules summarized in Section~\ref{sec:ibp}.
We describe the propagated state, the value and derivative updates for each
layer, and the resulting output bounds. Appendix~\ref{app:crown-derivation}
follows the same structure for CROWN.

\subsection{Propagated state and initialization}

For every layer $y = \mathcal{L}(z)$, suppose we are given
\begin{equation}
    \underline z(x_j)\le z(x_j)\le\overline z(x_j),
    \qquad
    \underline z'_j\le z'(x)\le\overline z'_j
    \quad\text{for every }x\in C_j.
\end{equation}
We derive IBP propagation rules for $y$ such that
\begin{equation}
    \underline y(x_j)\le y(x_j)\le\overline y(x_j),
    \qquad
    \underline y'_j\le y'(x)\le\overline y'_j
    \quad\text{for every }x\in C_j.
    \label{eq:ibp-appendix-state}
\end{equation}

At initialization, $z=a$: evaluate the input envelopes at the nodes and
use the prescribed input derivative bounds on each cell. The rules below
preserve these enclosures for every admissible input.

\subsection{Pointwise affine maps}

For \(y(x)=Wz(x)+b\), let
\(W^+=\max(W,0)\) and \(W^-=\min(W,0)\), componentwise. The value bounds are
\begin{align*}
    \underline y(x)
    &=W^+\underline z(x)+W^-\overline z(x)+b,\\
    \overline y(x)
    &=W^+\overline z(x)+W^-\underline z(x)+b.
\end{align*}
Because \(W\) and \(b\) are independent of \(x\), the corresponding
cellwise derivative bounds are
\begin{align*}
    \underline y'_j
    &=W^+\underline z'_j+W^-\overline z'_j,\\
    \overline y'_j
    &=W^+\overline z'_j+W^-\underline z'_j.
\end{align*}
The lifting, projection, and skip branches all use this rule.

\subsection{Pointwise activations}

Suppose \(\sigma\) is locally absolutely continuous and its derivative can
be enclosed over a bounded interval. For \(y(x)=\sigma(z(x))\), the value bounds are
\begin{align*}
    \underline y(x)
    &=\inf_{v\in[\underline z(x),\overline z(x)]}\sigma(v),\\
    \overline y(x)
    &=\sup_{v\in[\underline z(x),\overline z(x)]}\sigma(v).
\end{align*}
For a monotone activation, these reduce to
\(\underline y(x)=\sigma(\underline z(x))\) and
\(\overline y(x)=\sigma(\overline z(x))\).
Let \([\underline z_j^C,\overline z_j^C]\) enclose \(z(x)\) throughout
cell \(C_j\), and define
\begin{align*}
    \underline d_j
    &=\inf_{v\in[\underline z_j^C,\overline z_j^C]}\sigma'(v),&
    \overline d_j
    &=\sup_{v\in[\underline z_j^C,\overline z_j^C]}\sigma'(v).
\end{align*}
Since \(y'(x)=\sigma'(z(x))z'(x)\), the cellwise derivative bounds are
\begin{align*}
    \underline y'_j
    &=\min\{
      \underline d_j\underline z'_j,
      \underline d_j\overline z'_j,
      \overline d_j\underline z'_j,
      \overline d_j\overline z'_j\},\\
    \overline y'_j
    &=\max\{
      \underline d_j\underline z'_j,
      \underline d_j\overline z'_j,
      \overline d_j\underline z'_j,
      \overline d_j\overline z'_j\}.
\end{align*}

\subsection{Normalization}
\label{app:ibp-normalization}

Normalization statistics depend on the uncertain hidden field and must
therefore be enclosed along with its values. Consider
\[
    y(x)=\gamma\frac{z(x)-\mu(z)}{\sqrt{v(z)+\eta}}+\beta,
    \qquad \eta>0,
\]
where $\mu$ and $v$ are the mean and population variance over a fixed
normalization group. The averaging weights are nonnegative and sum to one;
for native-grid statistics, they are the model's prescribed discrete weights.
Suppose $|z-\widetilde z|\le r$ throughout this group and at every output
location being certified. Define
\[
    \widetilde\mu=\mu(\widetilde z),\qquad
    a=\sqrt{v(\widetilde z)+\eta},\qquad
    \widetilde y(x)=\gamma\frac{\widetilde z(x)-\widetilde\mu}{a}+\beta.
\]
Averaging the input error gives $|\mu(z)-\widetilde\mu|\le r$, so
\[
    |(z(x)-\mu(z))-(\widetilde z(x)-\widetilde\mu)|\le 2r.
\]
To bound the denominator, let $Pz=z-\mu(z)$ and let $\|\cdot\|_\mu$
be the root-mean-square norm under the same averaging weights.
Centering is an orthogonal projection in this norm. The reverse triangle
inequality applied to the augmented vectors $(Pz,\sqrt\eta)$ and
$(P\widetilde z,\sqrt\eta)$ therefore gives
\[
    \left|\sqrt{v(z)+\eta}-a\right|
    \le \|P(z-\widetilde z)\|_\mu
    \le \|z-\widetilde z\|_\mu\le r.
\]
Consequently, $s=\sqrt{v(z)+\eta}$ satisfies
$b=\max\{a-r,\sqrt\eta\}\le s$. Decomposing the normalized difference as
\[
    \frac{z-\mu(z)}{s}-\frac{\widetilde z-\widetilde\mu}{a}
    =\frac{(z-\mu(z))-(\widetilde z-\widetilde\mu)}{s}
    +(\widetilde z-\widetilde\mu)\frac{a-s}{as}
\]
yields the pointwise enclosure
\begin{equation}
    |y(x)-\widetilde y(x)|
    \le |\gamma|\left(
        \frac{2r}{b}
        +\frac{|\widetilde z(x)-\widetilde\mu|\,r}{ab}
    \right).
    \label{eq:ibp-normalization}
\end{equation}
A uniform output radius follows by replacing
$|\widetilde z(x)-\widetilde\mu|$ with a certified supremum over the output
domain. This rule is applied separately to each normalization group and
componentwise to fixed affine parameters $\gamma$ and $\beta$.

\subsection{Spectral convolution}

For
\begin{equation}
    y(x)=b+\int_\Omega\kappa(x-s)z(s)\,ds,
    \label{eq:ibp-spectral-layer}
\end{equation}
Componentwise sign splitting gives
\begin{equation}
\begin{aligned}
    \overline y(x)
    &=b+\int_\Omega
      \kappa^+(x-s)\overline z(s)
      +\kappa^-(x-s)\underline z(s)ds,\\
    \underline y(x)
    &=b+\int_\Omega
      \kappa^+(x-s)\underline z(s)
      +\kappa^-(x-s)\overline z(s)ds.
\end{aligned}
\label{eq:ibp-spectral-value}
\end{equation}
The two-point Gauss rule of Appendix~\ref{app:quad-constants}, with nodes \(x_{j,k}\) and node enclosures \(\underline z_{j,k}\), \(\overline z_{j,k}\) from \eqref{eq:gauss-node-enclosure}, converts these continuum bounds into the computable
enclosures
\begin{equation}
\begin{aligned}
    y(x)&\le b+\frac h2\sum_{j=0}^{Q-1}\sum_{k=1}^2
      \left[\kappa^+(x-x_{j,k})\overline z_{j,k}
      +\kappa^-(x-x_{j,k})\underline z_{j,k}\right]+\eta,\\
    y(x)&\ge b+\frac h2\sum_{j=0}^{Q-1}\sum_{k=1}^2
      \left[\kappa^+(x-x_{j,k})\underline z_{j,k}
      +\kappa^-(x-x_{j,k})\overline z_{j,k}\right]-\eta.
\end{aligned}
\label{eq:ibp-spectral-value-quadrature}
\end{equation}
For the 1st-derivative bounds, differentiating \eqref{eq:ibp-spectral-layer} with respect to \(x\) and
applying the same sign split gives
\begin{equation}
\begin{aligned}
    \overline y'(x)
    &=\int_\Omega
      \kappa'^+(x-s)\overline z(s)
      +\kappa'^-(x-s)\underline z(s)ds,\\
    \underline y'(x)
    &=\int_\Omega
      \kappa'^+(x-s)\underline z(s)
      +\kappa'^-(x-s)\overline z(s)ds.
\end{aligned}
\label{eq:ibp-appendix-derivative}
\end{equation}
Likewise, the differentiated integral is enclosed by
\begin{equation}
\begin{aligned}
    y'(x)&\le \frac h2\sum_{j=0}^{Q-1}\sum_{k=1}^2
      \left[\kappa'^+(x-x_{j,k})\overline z_{j,k}
      +\kappa'^-(x-x_{j,k})\underline z_{j,k}\right]+\eta',\\
    y'(x)&\ge \frac h2\sum_{j=0}^{Q-1}\sum_{k=1}^2
      \left[\kappa'^+(x-x_{j,k})\underline z_{j,k}
      +\kappa'^-(x-x_{j,k})\overline z_{j,k}\right]-\eta'.
\end{aligned}
\label{eq:ibp-spectral-derivative-quadrature}
\end{equation}
In the implementation, \eqref{eq:ibp-spectral-derivative-quadrature} is first
evaluated at the target node $x_q$. To obtain the cellwise derivative interval
required by \eqref{eq:ibp-appendix-state}, let
$S_{\kappa'',qj}\geq\sup_{x\in C_q,\,s\in C_j}|\kappa''(x-s)|$ and let
$S_{z,j}\geq\sup_{s\in C_j}|z(s)|$. We add the target-cell correction
\begin{equation*}
    \eta_{q}^{\prime,\mathrm{cell}}
    =h^2\sum_{j=0}^{Q-1}S_{\kappa'',qj}S_{z,j},
\end{equation*}
subtracting it from the nodal lower bound and adding it to the nodal upper
bound. Since $|x-x_q|\leq h$ for $x\in C_q$, the resulting interval encloses
$y'(x)$ throughout $C_q$.

\subsection{Output bounds}

After the final layer, the propagated intervals directly enclose the output
at the quadrature nodes. With
$M_{y,j}=\max\{|\underline y'_j|,|\overline y'_j|\}$, the extension rule
\eqref{extend} gives, for every $x\in C_j$,
\begin{equation*}
    \underline y(x_j)-M_{y,j}(x-x_j)
    \le y(x)\le
    \overline y(x_j)+M_{y,j}(x-x_j).
\end{equation*}
These bounds hold for every input in $\mathcal A$.

\section{Function-space CROWN propagation}
\label{app:crown-derivation}

This section gives the rules summarized in \secref{sec:crown}, following
the same organization as Appendix~\ref{app:ibp-rules}. We propagate affine
functionals backward, together with the coefficient derivative intervals
needed for validated integration.

\subsection{Propagated state and initialization}

We use
column vectors for channel dimensions and matrix-valued kernels. Fix an
output location \(x\in\Omega\). At any intermediate function
\(z:\Omega\to\mathbb{R}^{m}\), a CROWN bound has the affine-functional form
\begin{equation}
    \mathcal{B}_x[z]
    =d(x)+p(x)^\top z(x)
    +\int_\Omega c(x,s)^\top z(s)\,ds,
    \label{eq:crown-functional}
\end{equation}
where \(p(x)\in\mathbb{R}^{m}\) is the coefficient of the value at \(x\),
\(c(x,s)\in\mathbb{R}^{m}\) is the coefficient density over the domain, and
\(d(x)\in\mathbb{R}\) is the accumulated constant. Separate copies of
\((d,p,c)\) are propagated for the upper and lower bounds. Backpropagation
starts with \(\mathcal{B}_x[z^{(n)}]=z^{(n)}(x)\), or
\begin{equation*}
    d(x)=0,\qquad p(x)=1,\qquad c(x,s)=0.
\end{equation*}
Each backward step substitutes a sound affine representation or relaxation
of the preceding layer into \eqref{eq:crown-functional}.

Validated quadrature in the backward pass requires bounds on the variation of
the coefficient density with respect to its integration variable. For a fixed
output location \(x\) and each cell \(C_j=[x_j,x_j+h)\), we maintain the
componentwise enclosures
\begin{equation*}
    \underline c_j\le c(x,x_j)\le\overline c_j,
    \qquad
    \underline c'_j\le \partial_s c(x,s)\le\overline c'_j
    \quad (s\in C_j).
\end{equation*}
They imply
\begin{equation*}
    \underline c_j^C=\underline c_j-hM_{c,j},\qquad
    \overline c_j^C=\overline c_j+hM_{c,j},\qquad
    M_{c,j}=\max\{|\underline c'_j|,|\overline c'_j|\},
\end{equation*}
where all operations are componentwise. At initialization, \(c=0\), so
\(\underline c'_j=\overline c'_j=0\).

\subsection{Pointwise affine maps}
Let a pointwise layer be
\begin{equation*}
    y(s)=Wz(s)+b,
\end{equation*}
with \(W\in\mathbb{R}^{n\times m}\) and \(b\in\mathbb{R}^{n}\). Substitution
into an affine functional of \(y\) gives
\begin{align*}
    \mathcal{B}_x[y]
    ={}&d+p^\top\bigl(Wz(x)+b\bigr)
       +\int_\Omega c(s)^\top\bigl(Wz(s)+b\bigr)\,ds \\
    ={}&d+p^\top b+\int_\Omega c(s)^\top b\,ds
       +(W^\top p)^\top z(x)
       +\int_\Omega (W^\top c(s))^\top z(s)\,ds.
\end{align*}
Here and below, dependence on the fixed output location \(x\) is suppressed
inside derivations. Matching coefficients with
\eqref{eq:crown-functional} yields
\begin{equation*}
    \begin{aligned}
        \widetilde d&=d+p^\top b+\int_\Omega c(s)^\top b\,ds,\\
        \widetilde p&=W^\top p,\\
        \widetilde c(s)&=W^\top c(s).
    \end{aligned}
\end{equation*}
The transpose appears because CROWN coefficients are covectors acting on the
layer output. This is the orientation used by the implementation's affine
backward step.

\paragraph{Derivative bounds.}
For a pointwise affine map, differentiating
\(\widetilde c=W^\top c\) gives
\(\partial_s\widetilde c=W^\top\partial_s c\). Therefore, with
\(W^+=\max(W,0)\) and \(W^-=\min(W,0)\),
\begin{equation}
    \begin{aligned}
    \underline{\widetilde c}'_j
      &=(W^+)^\top\underline c'_j+(W^-)^\top\overline c'_j,\\
    \overline{\widetilde c}'_j
      &=(W^+)^\top\overline c'_j+(W^-)^\top\underline c'_j.
    \end{aligned}
    \label{eq:crown-affine-derivative-rule}
\end{equation}
Derivative intervals from additive branches are added endpointwise.

\subsection{Pointwise activations}

Suppose interval propagation has established
\(\underline z_i(s)\le z_i(s)\le\overline z_i(s)\). For an elementwise
activation \(\sigma\), choose valid affine relaxations
\begin{equation*}
    \alpha_i^L(s)z_i(s)+\beta_i^L(s)
    \le \sigma(z_i(s))
    \le \alpha_i^U(s)z_i(s)+\beta_i^U(s).
\end{equation*}
For a coefficient \(q_i\), multiplication by a negative value reverses the
inequality. Therefore,
\begin{align}
    q_i\sigma(z_i)
    &\le
    \bigl(q_i^+\alpha_i^U+q_i^-\alpha_i^L\bigr)z_i
    +q_i^+\beta_i^U+q_i^-\beta_i^L,
    \label{eq:activation-upper-sub}\\
    q_i\sigma(z_i)
    &\ge
    \bigl(q_i^+\alpha_i^L+q_i^-\alpha_i^U\bigr)z_i
    +q_i^+\beta_i^L+q_i^-\beta_i^U.
    \nonumber
\end{align}
where \(q_i^+=\max(q_i,0)\) and \(q_i^-=\min(q_i,0)\). Define
\begin{align*}
    A^U(q,s)&=q^+\odot\alpha^U(s)+q^-\odot\alpha^L(s),&
    B^U(q,s)&=q^+\odot\beta^U(s)+q^-\odot\beta^L(s),\\
    A^L(q,s)&=q^+\odot\alpha^L(s)+q^-\odot\alpha^U(s),&
    B^L(q,s)&=q^+\odot\beta^L(s)+q^-\odot\beta^U(s).
\end{align*}
Applying \eqref{eq:activation-upper-sub} to both coefficient terms in
\eqref{eq:crown-functional} gives
\begin{equation*}
    \begin{aligned}
        \widetilde d
        &=d+\mathbf{1}^\top B^U(p,x)
          +\int_\Omega\mathbf{1}^\top B^U(c(s),s)\,ds,\\
        \widetilde p&=A^U(p,x),\\
        \widetilde c(s)&=A^U(c(s),s).
    \end{aligned}
\end{equation*}
The lower-bound rule replaces \(A^U,B^U\) by \(A^L,B^L\). Thus sign
splitting is performed pointwise for both the local and kernel coefficients.

For a general activation, the implementation uses any
\((\alpha^L,\beta^L)\) and \((\alpha^U,\beta^U)\) satisfying the affine
relaxations above over the corresponding preactivation interval. These
parameters are held constant within each quadrature cell and computed from
the cellwise preactivation bounds. Convex or concave intervals use tangent
and secant bounds, while intervals crossing a change in curvature use valid
tangent--secant relaxations or are partitioned into smaller cells.

\paragraph{Derivative bounds.}
For an activation backward step, consider one coefficient component on one
cell. Both the upper and lower rules can be written
\begin{equation*}
    \widetilde c=\alpha_+c^++\alpha_-c^-,
\end{equation*}
where \((\alpha_+,\alpha_-)=(\alpha^U,\alpha^L)\) for an upper
functional and \((\alpha^L,\alpha^U)\) for a lower functional. The slopes
are held constant within each quadrature cell. For a scalar \(a\) and an
interval \(I=[\underline v,\overline v]\), define
\begin{equation*}
    aI=[\min\{a\underline v,a\overline v\},
        \max\{a\underline v,a\overline v\}].
\end{equation*}
The derivative enclosure is
\begin{equation*}
 [\underline{\widetilde c}'_j,\overline{\widetilde c}'_j]
 =
 \begin{cases}
   \alpha_+[\underline c'_j,\overline c'_j],
       &\underline c_j^C\ge0,\\
   \alpha_-[\underline c'_j,\overline c'_j],
       &\overline c_j^C\le0,\\
   \operatorname{hull}\!\left(
      \alpha_+[\underline c'_j,\overline c'_j],
      \alpha_-[\underline c'_j,\overline c'_j]\right),
       &\underline c_j^C<0<\overline c_j^C.
 \end{cases}
\end{equation*}
The last case encloses the change of slope where \(c\) crosses zero. The
activation-intercept integrand
\(\beta_+c^++\beta_-c^-\) uses the same rule with
\((\beta_+,\beta_-)\) in place of \((\alpha_+,\alpha_-)\).

\subsection{Spectral convolution}

Write the real-valued continuum spectral layer as
\begin{equation*}
    y(r)=b+\int_\Omega K(r,t)z(t)\,dt,
\end{equation*}
where \(K(r,t)\in\mathbb{R}^{n\times m}\). For an FNO, \(K\) is induced by
the retained Fourier modes. The local coefficient contributes
\begin{equation*}
    p^\top y(x)
    =p^\top b
     +\int_\Omega\bigl(K(x,t)^\top p\bigr)^\top z(t)\,dt.
\end{equation*}
For the kernel coefficient, Fubini's theorem gives
\begin{align*}
    \int_\Omega c(s)^\top y(s)\,ds
    ={}&\int_\Omega c(s)^\top b\,ds \\
      &+\int_\Omega
        \left(\int_\Omega K(s,t)^\top c(s)\,ds\right)^\top
        z(t)\,dt.
\end{align*}
Combining these two terms yields
\begin{equation}
    \begin{aligned}
        \widetilde d
        &=d+p^\top b+\int_\Omega c(s)^\top b\,ds,\\
        \widetilde p&=0,\\
        \widetilde c(t)
        &=K(x,t)^\top p
          +\int_\Omega K(s,t)^\top c(s)\,ds.
    \end{aligned}
    \label{eq:crown-spectral-rule}
\end{equation}
The first term in \(\widetilde c\) is the localized kernel generated by
\(p\); the second is the adjoint convolution applied to the existing
coefficient field. These are the two fields added by the spectral backward
implementation. No sign split is needed at the mathematical level because
the layer is linear; interval sign splitting is used only to enclose finite
representations of the coefficient field.

An FNO preactivation combines pointwise and spectral branches,
\begin{equation*}
    y(r)=Wz(r)+b+\int_\Omega K(r,t)z(t)\,dt.
\end{equation*}
Since the branches add, their backward coefficients add:
\begin{equation}
    \begin{aligned}
        \widetilde d
        &=d+p^\top b+\int_\Omega c(s)^\top b\,ds,\\
        \widetilde p&=W^\top p,\\
        \widetilde c(t)
        &=W^\top c(t)+K(x,t)^\top p
          +\int_\Omega K(s,t)^\top c(s)\,ds.
    \end{aligned}
    \label{eq:crown-fno-rule}
\end{equation}
The implementation propagates the spectral and skip branches from zero-bias
copies of the shared state, adds their coefficients, and restores the shared
bias once. This is algebraically \eqref{eq:crown-fno-rule} and avoids
double-counting \(d\).

\paragraph{Derivative bounds.}
For the spectral backward rule in \eqref{eq:crown-spectral-rule}, direct
differentiation with respect to the new source variable \(t\) gives
\begin{equation*}
    \partial_t\widetilde c(t)
    =-K'(x-t)^\top p
     -\int_\Omega K'(s-t)^\top c(s)\,ds.
\end{equation*}
This identity is enclosed by corrected quadrature. At a target node \(t_q\), the componentwise quadrature correction is
\begin{equation*}
    \eta_{c',q}^{\mathrm{src}}
    =d\,h^2\sum_j
      \left(S_{K'',qj}S_{c,j}+S_{K',qj}M_{c,j}\right).
\end{equation*}
Here \(S_{K',qj}\) and \(S_{K'',qj}\) bound the absolute entries of the
kernel derivatives on source cell \(C_j\), with sums over contracted channels
understood. Finally, to cover every \(t\in[t_q,t_q+h)\), rather than only
\(t_q\), we add
\begin{equation*}
    \eta_{c',q}^{\mathrm{cell}}
    =h\left(S_{K'',xq}|p|
      +h\sum_j S_{K'',qj}S_{c,j}\right).
\end{equation*}
The source integral is evaluated with the two-point Gauss rule at the nodes \(x_{j,k}=x_j+t_kh\), using the node enclosures
\(\underline c_{j,k}=\underline c_j-t_khM_{c,j}\) and \(\overline c_{j,k}=\overline c_j+t_khM_{c,j}\) as in \eqref{eq:gauss-node-enclosure}.
The spectral derivative bounds stored for the target cell are therefore
\begin{equation}
    \begin{aligned}
    \underline{\widetilde c}'_q
      &=-K'(x-t_q)^\top p\\
      &\quad+\frac h2\sum_j\sum_{k=1}^2\left(
          [-K'(x_{j,k}-t_q)^\top]^+\underline c_{j,k}
         +[-K'(x_{j,k}-t_q)^\top]^-\overline c_{j,k}\right)
        -\eta_{c',q}^{\mathrm{src}}
        -\eta_{c',q}^{\mathrm{cell}},\\
    \overline{\widetilde c}'_q
      &=-K'(x-t_q)^\top p\\
      &\quad+\frac h2\sum_j\sum_{k=1}^2\left(
          [-K'(x_{j,k}-t_q)^\top]^+\overline c_{j,k}
         +[-K'(x_{j,k}-t_q)^\top]^-\underline c_{j,k}\right)
        +\eta_{c',q}^{\mathrm{src}}
        +\eta_{c',q}^{\mathrm{cell}}.
    \end{aligned}
    \label{eq:crown-spectral-derivative-bound}
\end{equation}
For the combined FNO layer, the interval from
\eqref{eq:crown-affine-derivative-rule} is added to
\eqref{eq:crown-spectral-derivative-bound}.

\subsection{Output bounds}

\paragraph{Concretization at the input set.}

After all layers have been eliminated, the propagated functional acts on
the input \(a\). For the pointwise box
\begin{equation*}
    \mathcal{A}
    =\{a:\underline a(s)\le a(s)\le\overline a(s)
      \text{ for all }s\in\Omega\},
\end{equation*}
the upper certificate is
\begin{equation*}
    \begin{aligned}
    \overline y(x)=d
      &+p^+(x)^\top\overline a(x)
       +p^-(x)^\top\underline a(x)\\
      &+\int_\Omega
       \left(c^+(x,s)^\top\overline a(s)
       +c^-(x,s)^\top\underline a(s)\right)ds .
    \end{aligned}
\end{equation*}
The lower certificate exchanges \(\underline a\) and \(\overline a\):
\begin{equation*}
    \begin{aligned}
    \underline y(x)=d
      &+p^+(x)^\top\underline a(x)
       +p^-(x)^\top\overline a(x)\\
      &+\int_\Omega
       \left(c^+(x,s)^\top\underline a(s)
       +c^-(x,s)^\top\overline a(s)\right)ds .
    \end{aligned}
\end{equation*}
Together, they are certified lower and upper bounds on the output at location \(x\):
\begin{align*}
    \underline{u}(x)\leq u_\theta(x)\leq \overline{u}(x)\quad\text{for every }a\in\mathcal{A},
\end{align*}
where \(u_\theta=\mathcal{G}_\theta(a)\).

At quadrature nodes, these bounds extend to the whole cell using
\eqref{extend} and the output derivative bounds from forward IBP.

\section{Two-dimensional cell and perturbation bounds}
\label{app:2d-linearization}

In one dimension, we can divide the domain into millions of tiny intervals ($Q \ge 10^6$). Each cell is so narrow that approximating the function inside it by a simple constant interval $[v_{\min}, v_{\max}]$ introduces negligible error. In two dimensions and in space--time, however, computational constraints limit the grid to much coarser cells ($Q \approx 64$--$512$). At the same time, computing backward bounds like function-space CROWN becomes too expensive, scaling as $\mathcal O(Q^{2d}\log Q)$.

On these coarser grids, plain interval arithmetic fails because it throws away all information about which way the function is sloping. When two terms slope in opposite directions—such as when adding the horizontal and vertical second derivatives in a Laplacian ($u_{xx} + u_{yy}$), or adding a skip connection—plain intervals add their widths together instead of canceling out.

To keep the certification tight at fast $\mathcal O(Q^d\log Q)$ forward cost, we track how functions slope and vary across cells and perturbations:
\begin{enumerate}[leftmargin=*]
    \item \textbf{Tracking slopes across 2D cells for ReLU networks (Section~\ref{app:bilinear-unperturbed}):} Instead of enclosing each cell with flat horizontal upper and lower bounds, we fit a surface across the four cell corners (bilinear interpolation). This captures the tilt of the function inside the cell without inflating the bounds.
    \item \textbf{Tracking linear responses to input perturbations (Section~\ref{app:linearized-perturbation}):} When the input function is perturbed by noise of size $\varepsilon$, we compute how the network responds linearly to that perturbation through the Fourier layers. Opposite signs in the perturbation can then cancel out, leaving only a tiny quadratic remainder $\mathcal O(\varepsilon^2)$ to be bounded as an interval.
    \item \textbf{Tracking cell slopes for derivatives and PDE residuals (Section~\ref{app:taylor-cells}):} For smooth ($\tanh$ or GELU) networks and differential equations, we represent the solution and its derivatives inside each cell as a center value plus a linear slope. Because the slopes keep their signs, opposite gradients cancel out when evaluating PDE residuals.
\end{enumerate}

\subsection{Bilinear cell bounds}
\label{app:bilinear-unperturbed}

On each cell $C=[x_0,x_0+h_x]\times[y_0,y_0+h_y]$, we represent the lower
and upper envelopes by bilinear surfaces. In local coordinates
$s=(x-x_0)/h_x$ and $t=(y-y_0)/h_y$, with $(s,t)\in[0,1]^2$, these satisfy
\begin{equation}
\begin{aligned}
&(1-s)(1-t)\,\underline z_{C,00}
+(1-s)t\,\underline z_{C,01}
+s(1-t)\,\underline z_{C,10}
+st\,\underline z_{C,11}
\\
&\qquad\le z(x,y)\le
(1-s)(1-t)\,\overline z_{C,00}
+(1-s)t\,\overline z_{C,01}
+s(1-t)\,\overline z_{C,10}
+st\,\overline z_{C,11}.
\end{aligned}
\label{eq:2d-bilinear-envelope}
\end{equation}
For comparison, our derivative-based construction stores only the
interval $[\underline z_{C,00},\overline z_{C,00}]$ at the lower-left corner,
together with coordinate-wise derivative bounds $M_{x,C}$ and $M_{y,C}$, and
extends this interval across the cell using
\begin{equation}
    \underline z_{C,00}-M_{x,C}(x-x_0)-M_{y,C}(y-y_0)
    \le z(x,y)\le
    \overline z_{C,00}+M_{x,C}(x-x_0)+M_{y,C}(y-y_0).
    \label{eq:2d-derivative-envelope}
\end{equation}
The bilinear bound instead retains a separate
value at each of the four corners and interpolates those values
across the cell. It therefore preserves spatial variation that the
single-corner envelope collapses, without propagating the two
derivative bounds $M_{x,C}$ and $M_{y,C}$ through every layer.
\begin{figure}[t]
    \centering
    \includegraphics[width=0.75\linewidth]{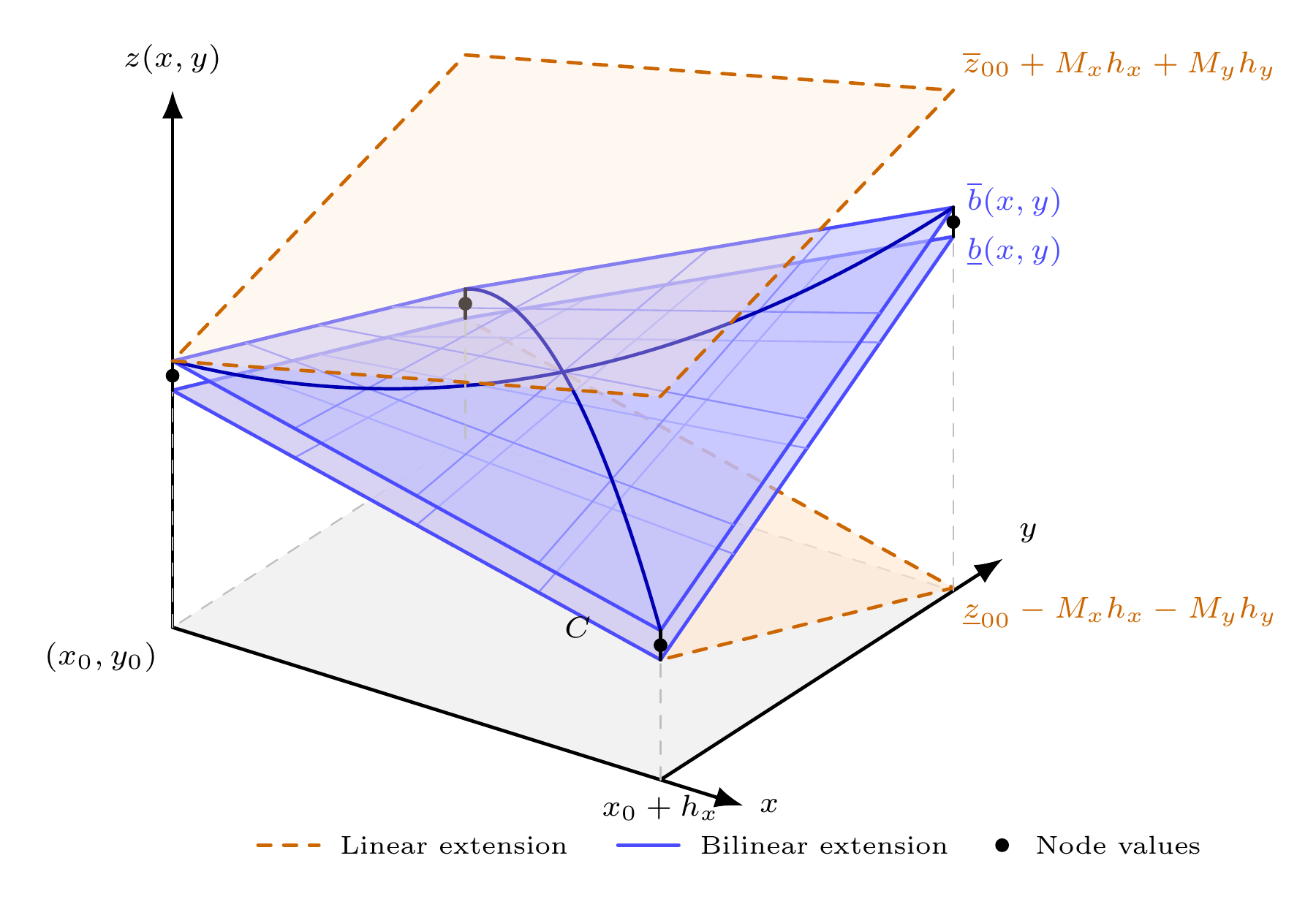}
    \caption{Illustration of bilinear and linear extension.}
    \label{fig:bilinear}
\end{figure}

\paragraph{Input bounds.}
Let $\mathcal I_C$ denote bilinear interpolation of the four corner values on
$C$, and write $\xi_{C,ij}=(x_0+ih_x,y_0+jh_y)$. The functions being bounded
need not themselves be bilinear. For any twice differentiable
unperturbed input $a_0$, the interpolation error satisfies
\begin{equation}
    \|a_0-\mathcal I_C a_0\|_{\infty,C}
    \le \frac{h_x^2}{8}\|\partial_{xx}a_0\|_{\infty,C}
       +\frac{h_y^2}{8}\|\partial_{yy}a_0\|_{\infty,C}
    =: \delta_a.
    \label{eq:2d-input-error}
\end{equation}
For the 2D Poisson forcing $a_0(x,y)=\sin(\pi x)\sin(\pi y)+\tfrac12\sin(2\pi x)\sin(2\pi y)$ on a $Q\times Q$ grid, $\|\partial_{xx}a_0\|_\infty,\|\partial_{yy}a_0\|_\infty\le3\pi^2$, which gives $\delta_a=3\pi^2/(4Q^2)$; for the Burgers initial condition $a_0(x)=\sin(2\pi x)+\tfrac12\cos(4\pi x)$, $\|a_0''\|_\infty\le12\pi^2$ gives $\delta_a=12\pi^2/(8Q^2)$.
We initialize the unperturbed corner coefficients as
\begin{equation}
    \underline a_{C,ij}=a_0(\xi_{C,ij})-\delta_a,
    \qquad
    \overline a_{C,ij}=a_0(\xi_{C,ij})+\delta_a.
    \label{eq:2d-input-corners}
\end{equation}
Only the known function $a_0$ is differentiated; no derivative constraint is imposed on the perturbation $a-a_0$.

\paragraph{Affine and activation propagation.}
All following inequalities act componentwise across channels. For a pointwise affine map $g=Wz+b$, let
$W^+=\max(W,0)$ and $W^-=\min(W,0)$. Each corner is updated by
\begin{align}
    \underline g_{C,ij}
        &=W^+\underline z_{C,ij}+W^-\overline z_{C,ij}+b,\nonumber\\
    \overline g_{C,ij}
        &=W^+\overline z_{C,ij}+W^-\underline z_{C,ij}+b.
    \label{eq:2d-affine-corners}
\end{align}
Because bilinear interpolation is a convex combination of corner values,
these updates bound $Wz(x,y)+b$ everywhere in the cell.

For $g=\sigma(z)$ with $\sigma(r)=\max(0,r)$, define the range of the
\emph{lower} corner coefficients and its maximum ReLU secant gap:
\begin{equation}
    m_C=\min_{i,j}\underline z_{C,ij},\qquad
    M_C=\max_{i,j}\underline z_{C,ij},\qquad
    \gamma_C=
    \begin{cases}
        -m_CM_C/(M_C-m_C),&m_C<0<M_C,\\
        0,&\text{otherwise}.
    \end{cases}
    \label{eq:2d-relu-gap}
\end{equation}
The propagated coefficients are
\begin{equation}
    \underline g_{C,ij}=\sigma(\underline z_{C,ij})-\gamma_C,
    \qquad
    \overline g_{C,ij}=\sigma(\overline z_{C,ij}).
    \label{eq:2d-relu-corners}
\end{equation}
Both envelopes remain valid because ReLU is convex and monotone, and $\gamma_C$ is the largest gap between ReLU and its secant over the corner range. Smooth activations such as $\tanh$ are bounded analogously using secant and tangent lines.

\paragraph{Spectral propagation.}
Write $\xi=(x,y)$ for an output point and $\eta$ for an integration point.
Decompose the incoming envelopes into a bilinear center and radius,
\begin{equation}
    \mu=\tfrac12(\underline z+\overline z),\qquad
    r=\tfrac12(\overline z-\underline z),\qquad |z-\mu|\le r.
    \label{eq:2d-center-radius}
\end{equation}
For the spectral map $g(\xi)=b+\int_\Omega K(\xi-\eta)z(\eta)\,d\eta$,
where $K:=\kappa_\ell$ is the matrix-valued kernel of layer $\ell$,
define its central output
\begin{equation}
    g_\mu(\xi)=b+\int_\Omega K(\xi-\eta)\mu(\eta)\,d\eta,
    \qquad
    |g(\xi)-g_\mu(\xi)|
    \le\int_\Omega |K(\xi-\eta)|r(\eta)\,d\eta.
    \label{eq:2d-spectral-radius}
\end{equation}
Matrix absolute values are entrywise, and multiplication sums over input
channels. Since $K$ is a finite Fourier series, $g_\mu$ can be evaluated
using the exact Fourier integrals of the piecewise bilinear center:
\begin{equation}
    \widehat\mu_k=\sum_C\int_C \mu(\eta)e^{-2\pi\mathrm{i}k\cdot\eta}\,d\eta,
    \qquad
    g_\mu(\xi)=b+\sum_{k\in\mathcal K}
        \widehat K_k\widehat\mu_k e^{2\pi\mathrm{i}k\cdot\xi}.
    \label{eq:2d-spectral-center}
\end{equation}
Here $\mathcal K$ contains the retained modes, including conjugate modes;
each cell integral separates into two one-dimensional integrals of a linear
polynomial times an exponential.

At each output node $\xi_q$, choose a matrix bound
$B_{qC}\ge\sup_{\eta\in C}|K(\xi_q-\eta)|$ entrywise. Since the integral
of a bilinear radius is the cell area times its mean corner value, a nodal
radius is
\begin{equation}
    \rho_q=\sum_C |C|B_{qC}r_C^{\mathrm{avg}},
    \qquad r_C^{\mathrm{avg}}=\tfrac14\sum_{i,j=0}^1 r_{C,ij},
    \qquad g_\mu(\xi_q)-\rho_q\le g(\xi_q)\le g_\mu(\xi_q)+\rho_q.
    \label{eq:2d-spectral-nodes}
\end{equation}
The bound $B_{qC}$ follows from the kernel's value and derivatives near the cell center, and the sums are evaluated by FFT convolution.
To extend these nodal values across an output cell, we add the bilinear interpolation error
\begin{equation}
    \delta_g=\frac{h_x^2H_x+h_y^2H_y}{8},
    \label{eq:2d-spectral-error}
\end{equation}
where $H_x,H_y$ bound the second derivatives of the spectral output and are computed directly from its retained Fourier modes. The new bilinear corner coefficients are consequently
\begin{equation}
    \underline g_{C,ij}=g_\mu(\xi_{C,ij})-\rho(\xi_{C,ij})-\delta_g,
    \qquad
    \overline g_{C,ij}=g_\mu(\xi_{C,ij})+\rho(\xi_{C,ij})+\delta_g,
    \label{eq:2d-spectral-corners}
\end{equation}
where $\rho(\xi_q)=\rho_q$. The correction bounds interpolation of the actual
spectral output $g$ between its nodes. A parallel pointwise skip branch is
bounded by \eqref{eq:2d-affine-corners}, and the two branches' corner
coefficients are added.

\subsection{Linearized perturbation bounds}
\label{app:linearized-perturbation}

To certify robustness under an input perturbation $h = a - a_0$ with $|h(\xi)|\le\varepsilon$, we decompose the perturbed operator output into its unperturbed value, a linear response, and a nonlinear remainder:
\[
\mathcal G_\theta(a_0+h)
=\mathcal G_\theta(a_0)+Th+\rho(h),
\]
where \(a_0\) is the unperturbed input, \(T\) is a linear integral operator describing the first-order response to \(h\), and \(\rho(h)\) collects the higher-order nonlinear remainder, which we bound separately.

At each activation layer, we split the perturbation change around the unperturbed preactivation $z$ into a linear slope $D$ and a remainder $r(z,e)$:
\[
\sigma(z+e)-\sigma(z)=De+r(z,e).
\]
Keeping the linear term \(De\) preserves sign cancellation across layers and channels, while bounding the remainder \(r(z,e)\) over the full perturbation range ensures a rigorous certificate rather than a local linear approximation.

Specifically, we first propagate the unperturbed input \(a_0\) with the bilinear rules above (\(\varepsilon=0\)) and fix one slope \(D\) per cell from the resulting unperturbed range. With these cellwise slopes fixed, the linear terms \(De\) compose through the affine and Fourier layers into a single linear operator \(Th(\xi)=p(\xi)h(\xi)+\int_\Omega K(\xi,\eta)h(\eta)\,d\eta\). On the unit domain, Cauchy--Schwarz bounds the linear response for any \(|h|\le\varepsilon\) by
\[
    |Th(\xi)|\le\varepsilon\bigl(|p(\xi)|+\|K(\xi,\cdot)\|_{L^2(\Omega)}\bigr),
\]
where \(\|K(\xi,\cdot)\|_{L^2(\Omega)}\) is computed in closed form from the exact cellwise Fourier moments of the piecewise-constant slopes (for space--time Burgers, where \(h(x)\) is shared across all times, we integrate \(K\) over time before taking the spatial \(L^2\) norm). The activation remainders \(r(z,e)\) are bounded on each cell from the worst-case deviation of \(\sigma\) from its chosen slope \(D\) and propagated through subsequent layers as ordinary interval radii, yielding an upper bound on \(|\rho(h)|\). Adding these bounds to the bilinear bound of \(\mathcal G_\theta(a_0)\) certifies the perturbed output over the entire domain.


\subsection{First-order Taylor cell bounds for higher-order derivatives}
\label{app:taylor-cells}

When certifying smooth ($\tanh$ or GELU) neural operators and their PDE residuals in two spatial dimensions or space--time, we must bound both the field $u_\theta$ and its higher-order partial derivatives (such as $u_x, u_y, u_{xx}, u_{yy}$ for 2D Poisson or $u_t, u_x, u_{xx}, u_{xxxx}$ for Burgers and Kuramoto--Sivashinsky) across every grid cell. If each derivative on a cell $[x_0-h_x,x_0+h_x]\times[y_0-h_y,y_0+h_y]$ were stored as a plain interval $[\underline v, \overline v]$, adding the spectral and skip branches or multiplying derivatives in the chain rule would add interval widths of size $\mathcal O(h_x + h_y)$ at every layer.

Instead, on each cell centered at $(x_0, y_0)$, we represent every intermediate function and partial derivative $f$ by its center value $c$, its linear slopes $(g_x, g_y)$ across the cell, and a higher-order remainder radius $r\ge 0$:
\begin{equation}
    f(x_0+\delta_x, y_0+\delta_y)
    \in c + g_x\delta_x + g_y\delta_y + [-r,\, r],
    \qquad |\delta_x|\le h_x,\quad |\delta_y|\le h_y.
    \label{eq:taylor-cell-form}
\end{equation}
Evaluated over the whole cell, \eqref{eq:taylor-cell-form} bounds $f$ within the interval $\bigl[c - (|g_x|h_x + |g_y|h_y + r),\, c + (|g_x|h_x + |g_y|h_y + r)\bigr]$. Because the slopes $(g_x, g_y)$ retain their signs throughout propagation, linear operations and nonlinear compositions preserve first-order cancellation:
\begin{itemize}[leftmargin=*]
    \item \textbf{Addition and linear layers:} Adding two branches adds their centers $c_1+c_2$ and slopes $(g_{1,x}+g_{2,x}, g_{1,y}+g_{2,y})$ directly so opposite slopes cancel, while adding only the remainders $r_1+r_2$. Similarly, a pointwise weight matrix $W$ acts directly on $c$ and $(g_x, g_y)$ and multiplies $r$ by $|W|$.
    \item \textbf{Products in the chain rule:} For products such as $(z_x)^2$ in $y_{xx} = \sigma''(z)(z_x)^2 + \sigma'(z)z_{xx}$, the product of two cell bounds keeps the exact linear part $c_1 g_2 + c_2 g_1$ and folds the quadratic cross-terms $\mathcal O(h^2)$ into the remainder $r$.
    \item \textbf{Smooth activations:} Composing \eqref{eq:taylor-cell-form} with $\sigma^{(k)}$ (for $k=0,1,\ldots$) by Taylor's theorem updates the center to $\sigma^{(k)}(c)$, scales the linear slope by $\sigma^{(k+1)}(c)$, and absorbs the second- and third-order Taylor terms into $r$, so the remainder $r$ scales as $\mathcal O(h^2)$ rather than $\mathcal O(h)$.
\end{itemize}
At spectral layers, the center $c$ and cell slopes $(g_x, g_y)$ of each partial derivative are synthesized directly from the certified Fourier coefficients of the preceding layer, while the Taylor remainder on the cell is bounded from the second and third spectral derivatives.
\subsection{Ablation of robustness bounds}
\label{app:robustness-ablation}

We test the separate and combined effects of bilinear cell bounds and linearized perturbation propagation on 2D Poisson and space--time Burgers. For each PDE, all four configurations use the same checkpoint, input and perturbation set. We use $2048\times2048$ cells for Poisson and $4096\times4096$ cells for Burgers. Disabling bilinear bounds replaces each operation's cell envelopes with constant intervals. Disabling linearized propagation uses ordinary interval propagation, which discards correlations between perturbations.

\begin{table}[H]
\centering
\small
\caption{Ablation of bilinear cell bounds and linearized perturbation propagation.}
\label{tab:robustness-ablation}
\begin{tabular}{lllrrrr}
\toprule
PDE & Bilinear & Linearized & Certified width & Gap & Time (s) & Memory (GiB)\\
\midrule
2D Poisson & Off & Off & $1.275\times10^{-1}$ & $2068.70$ & $1.31$ & $10.54$\\
& On & Off & $4.009\times10^{-2}$ & $650.51$ & $0.77$ & $15.60$\\
& Off & On & $8.725\times10^{-2}$ & $1415.56$ & $2.97$ & $13.44$\\
& On & On & $4.645\times10^{-4}$ & $7.54$ & $2.93$ & $18.51$\\
\midrule
Burgers & Off & Off & $5.008$ & $987.23$ & $9.81$ & $78.76$\\
& On & Off & $4.291$ & $846.03$ & $9.39$ & $116.26$\\
& Off & On & $3.750$ & $739.28$ & $64.15$ & $110.91$\\
& On & On & $2.465\times10^{-2}$ & $4.86$ & $63.89$ & $145.91$\\
\bottomrule
\end{tabular}
\end{table}

Table~\ref{tab:robustness-ablation} reports the maximum pointwise certified output width as well as gap against empirical samples. Either method alone reduces the width, but combining them gives a much larger improvement: approximately $274\times$ for Poisson and $203\times$ for Burgers compared with disabling both. This shows the benefit of tracking both variation within each cell and the effect of the shared input perturbation.
\section{Soundness proof of function-space certification}
\label{app:soundness}
\begin{theorem}[Soundness of adversarial robustness certification]
\label{thm:function-soundness}
Let \(\mathcal G_\theta\) be the continuum FNO and \(a_0\in\mathcal A\). Assume the inputs are
absolutely continuous with the prescribed value and derivative bounds. Then function-space IBP and CROWN, with the cellwise
extension in \eqref{extend}, produce bounds satisfying
\[
    \underline u(x)\le u_\theta(x)\le\overline u(x)
    \qquad\text{for every }a\in\mathcal A\text{ and }x\in\Omega,
\]
where \(u_\theta=\mathcal G_\theta(a)\), and the following inequality
\[
    \sup_{a\in\mathcal A}
    \|\mathcal G_\theta(a)-\mathcal G_\theta(a_0)\|_\infty\le\delta.
\]
\end{theorem}

\paragraph{Proof.}

The propagation rules in Appendix~\ref{app:ibp-rules} and~\ref{app:crown-derivation} establishes that for a layer
\(y=\mathcal L(z)\) with
\begin{equation*}
    \underline z(x_j)\le z(x_j)\le\overline z(x_j),
    \qquad
    \underline z'_j\le z'(x)\le\overline z'_j
    \quad\text{for every }x\in C_j.
\end{equation*}
It implies
\begin{align*}
    \underline y(x_j)\le y(x_j)\le\overline y(x_j)
    \qquad\text{for every }j.
\end{align*}

By definition of \(\mathcal A\),
\begin{align*}
    \underline a(x)\le a(x)\le\overline a(x)
    \qquad\text{for every }a\in\mathcal A\text{ and }x\in\Omega,
\end{align*}
and the prescribed derivative bounds hold throughout the input set.
Induction through the propagation steps gives
\begin{align*}
    \underline u(x_j)\le u_\theta(x_j)\le\overline u(x_j)
\end{align*}
for \(u_\theta=\mathcal G_\theta(a)\). To extend these to the full domain, let \(M_{u,j}\) be a
uniform bound on the absolute output derivative on \(C_j\), obtained from
forward propagation. Absolute continuity and \eqref{extend} give, for
\(x\in C_j\),
\begin{align*}
    \underline u(x)&=\underline u(x_j)-M_{u,j}(x-x_j),\\
    \overline u(x)&=\overline u(x_j)+M_{u,j}(x-x_j).
\end{align*}
Consequently,
\begin{align*}
    \underline u(x)\le u_\theta(x)\le\overline u(x)
    \qquad\text{for every }a\in\mathcal A\text{ and }x\in\Omega.
\end{align*}
Since \(a_0\in\mathcal A\), the unperturbed output lies in the same enclosure, and hence
\begin{align*}
    \sup_{a\in\mathcal A}
    \|\mathcal G_\theta(a)-\mathcal G_\theta(a_0)\|_\infty
    \le\max_{x\in\Omega}\big(\overline u(x)-\underline{u}(x)\big)\le\max_j \big(\overline u(x_j)-\underline{u}(x_j) + 2M_{u,j}h\big).
\end{align*}
We define
\begin{align*}
    \delta=\max_j \big(\overline u(x_j)-\underline{u}(x_j) + 2M_{u,j}h\big).
\end{align*}
This establishes the claimed robustness bound. The soundness of other fixed-input properties can be proven in a similar manner, by setting $\underline a(x)=a(x)=\overline a(x)$.

\section{PDE stability estimates}
\label{app:solution-stability}

A prediction can satisfy a differential equation approximately without being close to its exact solution. The missing ingredient is a \emph{stability estimate}: a bound on how much equation, boundary, and initial-condition errors can change the solution. This appendix derives the estimates used in Section~\ref{sec:solution-certification}, following the residual-based approach of \citet{ainsworth2000posteriori,mishra2023estimates}.

Throughout, we fix the input $a$, write $v=\mathcal G_\theta(a)$ for the predicted field and $u=\mathcal G(a)$ for the exact solution, and set $e=v-u$. Subtracting their equations gives an equation for $e$ driven by the computable residual $r$. We then bound $e$ using the maximum principle for Poisson, a Green's function for Helmholtz, and energy estimates for Burgers and Kuramoto--Sivashinsky.

\subsection{Poisson equation}
\label{app:poisson-stability}

\paragraph{One dimension.}
Consider the Dirichlet problem
\[
    -u_{xx}=a \quad\text{on }(0,L),
    \qquad u(0)=g(0),\quad u(L)=g(L).
\]
For the prediction $v$, define $r=-v_{xx}-a$ and certify
\[
    \|r\|_\infty\le\varepsilon_r,
    \qquad \max\{|v(0)-g(0)|,|v(L)-g(L)|\}\le\delta_b.
\]
The error satisfies $-e_{xx}=r$, with endpoint values bounded by $\delta_b$. To bound the interior error, set
\[
    \phi(x)=\frac{x(L-x)}2,
    \qquad -\phi_{xx}=1,
    \qquad 0\le\phi\le\frac{L^2}{8}.
\]
The functions $w_\pm=\delta_b+\varepsilon_r\phi\pm e$ are nonnegative at both endpoints and satisfy $-(w_\pm)_{xx}=\varepsilon_r\pm r\ge0$. By the maximum principle, they are nonnegative throughout the interval~\citep{evans2010partial,gilbarg2001elliptic}. Thus $|e|\le\delta_b+\varepsilon_r\phi$, giving
\begin{equation}
    \|v-u\|_\infty\le\delta_u=\delta_b+\frac{L^2}{8}\varepsilon_r.
    \label{eq:poisson-solution-bound}
\end{equation}

For an $L^2$ bound, split $e=z+h$, where $h$ is the linear function matching the endpoint errors. Then $z$ has zero endpoint values and $-z_{xx}=r$, while $\|h\|_2\le\sqrt L\,\delta_b$. Multiplying the equation for $z$ by $z$ and integrating gives
\[
    \frac{\pi^2}{L^2}\|z\|_2^2
    \le\|z_x\|_2^2=\int_0^L rz\,dx
    \le\|r\|_2\|z\|_2.
\]
The first inequality is the Poincar\'e inequality for functions vanishing at both endpoints. Hence, if $\|r\|_2\le\varepsilon_{r,2}$,
\begin{equation}
    \|v-u\|_2\le\sqrt L\,\delta_b+\frac{L^2}{\pi^2}\varepsilon_{r,2}.
    \label{eq:poisson-l2-bound}
\end{equation}
For the 1D Poisson certificate in Table~\ref{tab:solution-results}, the boundary bound is $\delta_b\le2.424\times10^{-3}$.

\paragraph{Two dimensions.}
On the rectangle $\Omega=(0,L_x)\times(0,L_y)$, consider
\[
    -u_{xx}-u_{yy}=a \quad\text{in }\Omega,
    \qquad u=g \quad\text{on }\partial\Omega.
\]
Define $r=-v_{xx}-v_{yy}-a$ and certify $\|r\|_\infty\le\varepsilon_r$ in the rectangle and $\|v-g\|_\infty\le\delta_b$ on its entire boundary. The error satisfies $-e_{xx}-e_{yy}=r$. If $L_x\le L_y$, choose $\phi(x,y)=x(L_x-x)/2$; otherwise choose $\phi(x,y)=y(L_y-y)/2$. In either case,
\[
    -\phi_{xx}-\phi_{yy}=1,
    \qquad 0\le\phi\le\frac{\min(L_x,L_y)^2}{8}.
\]
As above, $w_\pm=\delta_b+\varepsilon_r\phi\pm e$ are nonnegative on the boundary and satisfy $-(w_\pm)_{xx}-(w_\pm)_{yy}=\varepsilon_r\pm r\ge0$. Applying the maximum principle gives
\begin{equation}
    \|v-u\|_\infty
    \le\delta_b+\frac{\min(L_x,L_y)^2}{8}\varepsilon_r=:\delta_u.
    \label{eq:poisson-2d-bound}
\end{equation}
Both the unit interval and the unit square therefore have residual coefficient $1/8$. In two dimensions, the boundary certificate must cover all four edges, not only the corners.

For an $L^2$ bound, split $e=z+h$, where $h$ satisfies $h_{xx}+h_{yy}=0$ with the same boundary values as $e$. The maximum principle gives $\|h\|_2\le\sqrt{L_xL_y}\,\delta_b$. The remaining error has zero boundary values and satisfies $-z_{xx}-z_{yy}=r$. Multiplying by $z$ and integrating gives
\[
    \pi^2(L_x^{-2}+L_y^{-2})\|z\|_2^2
    \le\|z_x\|_2^2+\|z_y\|_2^2
    =\int_\Omega rz\,dx\,dy\le\|r\|_2\|z\|_2.
\]
Here the first inequality follows by applying the one-dimensional Poincar\'e inequality in each coordinate. Thus, if $\|r\|_2\le\varepsilon_{r,2}$,
\begin{equation}
    \|v-u\|_2\le\sqrt{L_xL_y}\,\delta_b
      +\frac{\varepsilon_{r,2}}{\pi^2(L_x^{-2}+L_y^{-2})}.
    \label{eq:poisson-2d-l2-bound}
\end{equation}
For the 2D Poisson certificate in Table~\ref{tab:solution-results}, the bound over all four edges is $\delta_b\le3.283\times10^{-3}$.

\subsection{Helmholtz equation}
\label{app:helmholtz-stability}

Consider the periodic problem
\[
    -u_{xx}-\kappa^2u=a \quad\text{on }(0,L),
    \qquad u(0)=u(L),\quad u_x(0)=u_x(L).
\]
We assume $\kappa>0$ and $\kappa L\notin2\pi\mathbb Z$. At the excluded values, a nonzero periodic sine or cosine solves the homogeneous equation, so a zero residual cannot determine the solution uniquely. For the prediction, define
\[
    r=-v_{xx}-\kappa^2v-a,\qquad
    d_0=v(0)-v(L),\qquad d_1=v_x(0)-v_x(L),
\]
and certify $\|r\|_\infty\le\varepsilon_r$, $|d_0|\le\delta_{b,0}$, and $|d_1|\le\delta_{b,1}$. The error satisfies $-e_{xx}-\kappa^2e=r$ and has the same endpoint jumps $d_0,d_1$.

First remove these jumps. With $\alpha=\kappa L/2$, the function
\[
    h(x)=\frac{(d_1/\kappa)\cos(\kappa(x-L/2))
                  -d_0\sin(\kappa(x-L/2))}{2\sin\alpha}
\]
satisfies $-h_{xx}-\kappa^2h=0$, $h(0)-h(L)=d_0$, and $h_x(0)-h_x(L)=d_1$. Consequently, $z=e-h$ is periodic in both value and derivative and satisfies the same residual equation.

The periodic Green's function describes the response to a unit point source. For this operator it is
\[
    K(x)=-\frac{\cos(\kappa(x-L/2))}{2\kappa\sin\alpha},
    \qquad 0\le x\le L,
\]
extended periodically. Its derivative has a jump of $-1$ at the source, as required by the leading term $-u_{xx}$. Convolution with the residual gives $z(x)=\int_0^L K(x-y)r(y)\,dy$, and therefore
\[
    \|z\|_\infty\le C_{\kappa,L}\varepsilon_r,
    \qquad C_{\kappa,L}=\int_0^L|K(x)|\,dx
    =\frac{\int_0^\alpha|\cos s|\,ds}{\kappa^2|\sin\alpha|}.
\]
Adding the bound on $h$ yields
\begin{equation}
    \|v-u\|_\infty
    \le C_{\kappa,L}\varepsilon_r
      +\frac{\delta_{b,0}}{2|\sin\alpha|}
      +\frac{\delta_{b,1}}{2\kappa|\sin\alpha|}
    =:\delta_u.
    \label{eq:helmholtz-solution-bound}
\end{equation}
The constants grow near resonance, explaining why a tight residual certificate can still give a large solution-error bound.

For an $L^2$ bound, each periodic Fourier mode has eigenvalue $(2\pi n/L)^2-\kappa^2$. Dividing the residual coefficients by these eigenvalues and applying Parseval's identity gives
\begin{equation}
    \|v-u\|_2
    \le \frac{\varepsilon_{r,2}}{\min_{n\in\mathbb Z}|(2\pi n/L)^2-\kappa^2|}
      +\sqrt L\left(
        \frac{\delta_{b,0}}{2|\sin\alpha|}
        +\frac{\delta_{b,1}}{2\kappa|\sin\alpha|}\right),
    \label{eq:helmholtz-l2-bound}
\end{equation}
where $\|r\|_2\le\varepsilon_{r,2}$ and the boundary contribution uses $\|h\|_2\le\sqrt L\|h\|_\infty$.
For the Helmholtz certificate in Table~\ref{tab:solution-results}, the boundary bounds are $\delta_{b,0}\le1.765\times10^{-11}$ and $\delta_{b,1}\le1.672\times10^{-9}$.

\subsection{Burgers equation}
\label{app:burgers-stability}

Consider the viscous Burgers equation on a periodic spatial interval,
\[
    u_t+uu_x-\nu u_{xx}=0,
    \qquad x\in(0,L),\quad t\in(0,T],
    \qquad u(x,0)=u_0(x),\quad \nu>0.
\]
Here the prediction $v(x,t)$ must be available throughout the time interval, with both its value and spatial derivative periodic. A prediction only at the final time does not supply the time derivative needed to certify the residual. Define
\[
    r=v_t+vv_x-\nu v_{xx},
    \qquad \|v(\cdot,0)-u_0\|_2\le\delta_0.
\]
Unlike the stationary estimates above, the solution bound must account for how this initial error and the residual accumulate over time.

Substituting $u=v-e$ gives the error equation
\[
    e_t+ve_x+v_xe-ee_x-\nu e_{xx}=r.
\]
Multiply by $e$ and integrate over space. Periodicity removes the boundary terms and gives $\int e^2e_x\,dx=0$, leaving the energy identity
\begin{equation}
    \frac12\frac{d}{dt}\|e\|_2^2
    =-\nu\|e_x\|_2^2-\frac12\int_0^L v_xe^2\,dx
      +\int_0^L re\,dx.
    \label{eq:burgers-energy}
\end{equation}
The first term reduces the error through viscosity; the second can increase it where the prediction has a negative slope. To retain both effects, define the growth coefficient
\begin{equation}
    \gamma(t)=\sup_{0\ne w\in H^1_{\mathrm{per}}(0,L)}
    \frac{-\nu\|w_x\|_2^2-\tfrac12\int_0^L v_x(x,t)w(x)^2\,dx}
         {\|w\|_2^2}.
    \label{eq:burgers-growth}
\end{equation}
Here $H^1_{\mathrm{per}}$ consists of periodic functions with square-integrable values and first derivatives. Thus $\gamma(t)$ is the largest possible error-growth rate from the first two terms of \eqref{eq:burgers-energy}; it depends on the predicted field, not the unknown solution. Constant test functions give $\gamma(t)\ge0$, since $\int_0^L v_x\,dx=0$.

Cauchy--Schwarz bounds the last term by $\|r\|_2\|e\|_2$, giving $\frac{d}{dt}\|e\|_2\le\gamma(t)\|e\|_2+\|r\|_2$ whenever $\|e\|_2>0$. The resulting integral bound also holds at zero error by continuity. Multiplying by the integrating factor $e^{-G(t)}$, where $G(t)=\int_0^t\gamma(s)\,ds$, and integrating yields the Gr\"onwall estimate~\citep{evans2010partial,mishra2023estimates}
\begin{equation}
    \|v(\cdot,t)-u(\cdot,t)\|_2
    \le e^{G(t)}\delta_0
      +\int_0^t e^{G(t)-G(s)}\|r(\cdot,s)\|_2\,ds.
    \label{eq:burgers-solution-bound}
\end{equation}
The two terms describe amplification of the initial error and accumulation of residual forcing, respectively.

To evaluate this bound, divide $[0,T]$ into time intervals $[t_j,t_{j+1}]$. On each entire interval, certify $\gamma(t)\le\overline\gamma_j$ and $\|r(\cdot,t)\|_2\le R_j$, with $\overline\gamma_j\ge0$. Starting from $q_0=\delta_0$, integrate the scalar equation with these constant upper bounds:
\begin{equation}
    q_{j+1}=e^{\overline\gamma_j\Delta t_j}q_j+
    \begin{cases}
      R_j\dfrac{e^{\overline\gamma_j\Delta t_j}-1}{\overline\gamma_j},
        &\overline\gamma_j>0,\\[6pt]
      R_j\Delta t_j,&\overline\gamma_j=0,
    \end{cases}
    \qquad \Delta t_j=t_{j+1}-t_j.
    \label{eq:burgers-recurrence}
\end{equation}
This envelope is nondecreasing and bounds the error between time nodes as well. With $t_N=T$, the final certificate is therefore
\begin{equation}
    \sup_{0\le t\le T}\|v(\cdot,t)-u(\cdot,t)\|_2\le q_N=:\delta_u.
    \label{eq:burgers-uniform-bound}
\end{equation}
The quantities $R_j$ are spatial $L^2$ bounds valid at every time in the interval, not averages over space and time. A pointwise residual bound $|r|\le\varepsilon_{r,j}$ gives $R_j=\sqrt L\,\varepsilon_{r,j}$; direct spatial integration of the certified squared residual can give a tighter bound.

It remains to bound $\gamma$ without discarding viscosity. On each time interval, let $\overline V_j(x)$ be a piecewise-constant upper bound on $V(x,t)=-v_x(x,t)/2$; replacing $V$ by $\overline V_j$ can only increase \eqref{eq:burgers-growth}. We split the periodic Fourier basis into retained modes $|k|\le K$ and omitted modes. On the retained modes, $a_j$ is a certified upper bound on the largest eigenvalue of the Galerkin matrix of $\nu\partial_{xx}+\overline V_j$. With $V_{\max,j}=\max_x\overline V_j$ and $V_{\min,j}=\min_x\overline V_j$, the omitted modes contribute at most $d_j=V_{\max,j}-\nu[2\pi(K+1)/L]^2$, and the coupling between the two groups is at most $c_j=(V_{\max,j}-V_{\min,j})/2$. The largest eigenvalue of $\left(\begin{smallmatrix}a_j&c_j\\c_j&d_j\end{smallmatrix}\right)$ then gives
\begin{equation}
    \overline\gamma_j=
    \max\!\left\{0,\min\!\left[
      V_{\max,j},\,
      \frac{a_j+d_j+\sqrt{(a_j-d_j)^2+4c_j^2}}{2}
    \right]\right\}.
    \label{eq:burgers-growth-bound}
\end{equation}
The bound $V_{\max,j}$ follows by dropping the viscosity term. This construction covers all Fourier modes, including the constant mode, so the error need not have zero mean. Substituting these growth bounds and the certified residual and initial errors into \eqref{eq:burgers-recurrence} completes the solution certificate.

For the Burgers certificate in Table~\ref{tab:solution-results}, $\delta_0\le2.316\times10^{-3}$ and spatial periodicity holds by construction. We use $\nu=0.01$, $L=T=1$, 4,096 spatial cells, 1,024 time intervals of width $1/1024$, and $K=128$. The interval growth bounds $\overline\gamma_j$ range from approximately $1.471$ to $9.649$, with $\sum_j\overline\gamma_j\Delta t_j\le4.653$.

\subsection{Kuramoto--Sivashinsky equation}
\label{app:ks-stability}

Consider the Kuramoto--Sivashinsky equation
\[
    u_t+uu_x+u_{xx}+\nu u_{xxxx}=0,
    \qquad x\in(0,L),\quad t\in(0,T],
    \qquad u(x,0)=u_0(x),\quad \nu>0.
\]
Assume that $u$ and the prediction $v$ are sufficiently smooth, and that their values and first three spatial derivatives agree at $x=0$ and $x=L$. Define
\[
    r=v_t+vv_x+v_{xx}+\nu v_{xxxx},
    \qquad \|v(\cdot,0)-u_0\|_2\le\delta_0.
\]
As for Burgers, substitute $u=v-e$, multiply the error equation by $e$, and integrate over space. Periodicity removes the boundary terms and the integral of $e^2e_x$, giving
\begin{equation}
    \frac12\frac{d}{dt}\|e\|_2^2
    =\|e_x\|_2^2-\nu\|e_{xx}\|_2^2
      -\frac12\int_0^L v_xe^2\,dx+\int_0^L re\,dx.
    \label{eq:ks-energy}
\end{equation}
The second spatial derivative can increase the error, while the fourth derivative reduces it. Integration by parts and Cauchy--Schwarz give
\[
    \|e_x\|_2^2=-\int_0^L ee_{xx}\,dx
    \le\|e\|_2\|e_{xx}\|_2
    \le\frac{1}{4\nu}\|e\|_2^2+\nu\|e_{xx}\|_2^2.
\]
Suppose we certify $v_x(x,t)\ge m$ throughout the domain and $\|r(\cdot,t)\|_2\le R$ at every time. Then
\[
    \frac{d}{dt}\|e\|_2\le\lambda\|e\|_2+R,
    \qquad \lambda=\frac{1}{4\nu}-\frac m2.
\]
The coefficient $\lambda$ is positive because a periodic function cannot have a strictly positive spatial derivative everywhere. Applying the same integrating-factor argument as for Burgers yields
\begin{equation}
    \|v(\cdot,t)-u(\cdot,t)\|_2
    \le e^{\lambda t}\delta_0+
       \frac{e^{\lambda t}-1}{\lambda}R,
    \qquad
    \delta_u=e^{\lambda T}\delta_0+
       \frac{e^{\lambda T}-1}{\lambda}R.
    \label{eq:ks-solution-bound}
\end{equation}
Thus $\delta_u$ bounds the solution error throughout $[0,T]$. A certified pointwise residual bound $|r|\le\varepsilon_r$ supplies $R=\sqrt L\,\varepsilon_r$.

For our KS experiments, $L=6\pi$ and $\nu=0.01$, so $1/(4\nu)=25$. Even before accounting for negative slopes of $v$, this estimate gives an amplification factor of at least $e^{20}$ over $T=0.8$. Combined with the large certified residuals, the estimate can therefore be very loose.

\section{Experimental and implementation details}
\label{app:experiment-details}

\subsection{Datasets and equations}

Table~\ref{tab:appendix-data-models} summarizes the benchmark equations, datasets, and neural operator architectures evaluated in the main text. Across all experiments, we use standard Fourier Neural Operator (FNO) architectures with linear skip connections and a channel expansion ratio of two in the lifting and projection layers. Because certifying a differential equation residual requires differentiating the model output twice (or up to four times for the Kuramoto--Sivashinsky equation), we use smooth hyperbolic tangent ($\tanh$) or Gaussian error linear unit (GELU) activations for residual and solution-error certification, and rectified linear unit (ReLU) activations where piecewise-linear robustness bounds are evaluated. All synthetic datasets are generated with a fixed random seed of zero.

\begin{table}[H]
\centering
\footnotesize
\setlength{\tabcolsep}{4pt}
\caption{Datasets and neural operator architectures across all six benchmarks. Models with ReLU activations are used for adversarial robustness certification; smooth models ($\tanh$ or GELU) are used when certifying differential equation residuals and solution errors.}
\label{tab:appendix-data-models}
\begin{adjustbox}{max width=\linewidth,center}
\begin{tabular}{llccccccc}
\toprule
Benchmark & Verification task & Domain & Boundary & Train / Val / Test & Grid $N$ & Layers / Ch. & Modes & Activation \\
\midrule
1D Poisson & Robustness & $[0,1]$ & Dirichlet & $720 / 80 / 200$ & $64$ & $1 / 8$ & $24$ & ReLU \\
 & Residual / solution & $[0,1]$ & Dirichlet & $720 / 80 / 200$ & $64$ & $1 / 4$ & $32$ & $\tanh$ \\
1D Helmholtz & Robustness & $[0,1)$ & Periodic & $800 / \text{---} / 200$ & $64$ & $1 / 4$ & $32$ & ReLU \\
 & Residual / solution & $[0,1)$ & Periodic & $800 / \text{---} / 200$ & $64$ & $1 / 4$ & $32$ & $\tanh$ \\
2D Poisson & Robustness & $[0,1]^2$ & Dirichlet & $4000 / 500 / 500$ & $32\times 32$ & $2 / 8$ & $12\times 12$ & ReLU \\
 & Residual / solution & $[0,1]^2$ & Dirichlet & $4000 / 500 / 500$ & $32\times 32$ & $2 / 8$ & $12\times 12$ & $\tanh$ \\
Burgers & All tasks & $[0,1)\times[0,1]$ & Periodic in $x$ & $720 / 80 / 200$ & $16\times 17$ & $2 / 16$ & $16\times 16$ & $\tanh$ \\
KS rollout & Multi-step residual & $[0,6\pi)\times[0,0.1]$ & Periodic in $x$ & $512 / 128 / 128$ & $32\times 128$ & $1 / 64$ & $16\times 33$ & GELU \\
SFNO (ERA5) & Regional maximum & Globe ($6\text{ h}$) & Spherical & Pretrained & $721\times 1440$ & $8 / 384$ & $239$ & GELU \\
\bottomrule
\end{tabular}
\end{adjustbox}
\end{table}

\paragraph{Stationary equations (1D Poisson, 1D Helmholtz, and 2D Poisson).}
For 1D Poisson and 1D Helmholtz, the input forcing functions are random trigonometric series containing frequencies $m=1,\ldots,15$, with independent Gaussian sine and cosine coefficients scaled by $\exp[-(\pi m\,0.2)^2]$ so that higher frequencies decay smoothly. Exact reference solutions are obtained via the one-dimensional zero-Dirichlet Green's function for 1D Poisson and by exact spectral division for 1D Helmholtz. In 1D Helmholtz, we set the wave number to $\kappa=\sqrt{(8\pi)^2-0.02}$, which places the equation just below its fourth resonant frequency ($m=4$) so that small input perturbations produce large output changes. For 2D Poisson on the unit square, forcing functions are random combinations of $\sin(\pi m x)\sin(\pi n y)$ for modes $m,n=1,\ldots,12$ with Gaussian coefficients of standard deviation $1/6$, paired with their exact analytical solutions. The 1D Poisson ReLU model and both 2D Poisson models append spatial grid coordinates to the input; the remaining 1D models take the forcing function alone.

\paragraph{Time-dependent and global benchmarks (Burgers, KS, and SFNO).}
The Burgers benchmark predicts the space--time velocity field over $x\in[0,1)$ and $t\in[0,1]$ with viscosity $\nu=0.01$ from two input channels (the initial velocity profile and the time coordinate) on a $16\times 17$ space--time grid. For the chaotic Kuramoto--Sivashinsky (KS) equation ($\nu=0.01$ on $[0,6\pi)$), the model maps the spatial state forward by $\tau=0.1$ per step on a $32\times 128$ temporal--spatial grid (padded to $34$ time points), starting from the smooth initial condition $u_0(x)=\sin(x/3)+\tfrac12\cos(2x/3)$. Reference KS training trajectories are generated with a $128$-node fourth-order exponential time-differencing Runge--Kutta solver. Finally, the global weather experiment uses the pretrained Spherical FNO (SFNO) from NVIDIA Earth2Studio~\citep{Earth2Studio_Contributors_NVIDIA_Earth2Studio_2024}, which maps $73$ atmospheric variables and four surface/solar auxiliary channels from the ERA5 reanalysis (15 July 2023, 18:00 UTC, represented by spherical harmonics up to degree $239$) to a six-hour global forecast.

\subsection{Training configuration}

Table~\ref{tab:appendix-training} lists the optimization settings used to train each PDE model. Most models are trained with the AdamW optimizer and cosine learning-rate decay, using relative $L^2$ error or a combination of Sobolev ($H^1$) and maximum pointwise ($L^\infty$) losses to promote smooth derivatives.

\begin{table}[H]
\centering
\footnotesize
\setlength{\tabcolsep}{5pt}
\caption{Training hyperparameters for the PDE benchmarks.}
\label{tab:appendix-training}
\begin{adjustbox}{max width=\linewidth,center}
\begin{tabular}{llclrrrr}
\toprule
Benchmark & Model & Training loss & Optimizer / schedule & Epochs & Batch size & Learning rate & Weight decay \\
\midrule
1D Poisson & ReLU & Relative $L^2$ & AdamW / cosine & $800$ & $32$ & $2\times10^{-3}$ & $10^{-6}$ \\
 & $\tanh$ & Relative $L^2$ + mean error & AdamW / cosine & $500$ & $32$ & $10^{-3}$ & $10^{-4}$ \\
1D Helmholtz & ReLU, $\tanh$ & $H^1 + 0.1 L^\infty$ & AdamW / cosine & $1000$ & $32$ & $10^{-3}$ & $10^{-4}$ \\
2D Poisson & ReLU, $\tanh$ & Relative $L^2 + 0.1 H^1 + 0.1 L^\infty$ & AdamW / cosine & $200 / 400$ & $32$ & $10^{-3}$ & $10^{-6}$ \\
Burgers & $\tanh$ & Trajectory relative $L^2$ & Adam / step decay & $500$ & $20$ & $10^{-3}$ & $0$ \\
KS rollout & GELU & Data + physics rollout MSE & AdamW / two-stage & $200+200$ & $8$ & $2\times 10^{-3}$ & $10^{-4}$ \\
\bottomrule
\end{tabular}
\end{adjustbox}
\end{table}

To ensure that models certified for PDE residuals have accurate higher-order derivatives, we include a small physics-informed penalty on the equation residual during a brief fine-tuning stage for the 2D Poisson, Burgers, and KS models. For the 1D Poisson ReLU model, we select the architecture with the lowest validation error among four candidates whose grid-level CROWN bound width remains within twice that of a smaller four-channel baseline.

\subsection{Test prediction errors}
\label{app:checkpoint-errors}

Table~\ref{tab:checkpoint-errors} reports the average prediction errors on held-out test sets at the native training resolution ($200$ test examples at $N=64$ for 1D Poisson and 1D Helmholtz, and $500$ examples at $32\times 32$ for 2D Poisson). We report both the relative $L^2$ error ($\|u_\theta - u\|_2 / \|u\|_2$) and the maximum pointwise error ($\|u_\theta - u\|_\infty$). For space--time Burgers, the trained $\tanh$ model achieves a mean validation relative $L^2$ error of $2.65\times 10^{-3}$ across the space--time trajectory and $2.37\times 10^{-3}$ on the nominal test profile in Table~\ref{tab:solution-results}.

\begin{table}[H]
\centering
\small
\caption{Mean test prediction errors at the native training resolution.}
\label{tab:checkpoint-errors}
\begin{adjustbox}{max width=\linewidth,center}
\begin{tabular}{llrr}
\toprule
Benchmark & Model & Relative $L^2$ error & Absolute $L^\infty$ error \\
\midrule
1D Poisson & ReLU (robustness) & $2.522\times10^{-2}$ & $8.287\times10^{-4}$ \\
 & $\tanh$ (residual / solution) & $5.831\times10^{-1}$ & $1.352\times10^{-2}$ \\
1D Helmholtz & ReLU (robustness) & $2.518\times10^{-4}$ & $5.780\times10^{-5}$ \\
 & $\tanh$ (residual / solution) & $1.028\times10^{-2}$ & $1.393\times10^{-3}$ \\
2D Poisson & ReLU (robustness) & $3.827\times10^{-2}$ & $8.086\times10^{-4}$ \\
 & $\tanh$ (residual / solution) & $4.995\times10^{-2}$ & $9.897\times10^{-4}$ \\
\bottomrule
\end{tabular}
\end{adjustbox}
\end{table}

\subsection{Certified input sets}
\label{app:certified-inputs}

Table~\ref{tab:certified-inputs} summarizes the reference inputs, perturbation radii, derivative bounds, and quadrature resolutions used in our certification experiments. For 1D Poisson, 1D Helmholtz, and space--time Burgers on $[0,L]$ with $L=1$, the unperturbed spatial input is
\begin{equation}
    a_0(x)=\sin\left(\frac{2\pi x}{L}\right)
    +\frac12\cos\left(\frac{4\pi x}{L}\right).
    \label{eq:certified-center}
\end{equation}
Under a pointwise perturbation of radius $\varepsilon=10^{-3}$, any admissible input $a(x)$ lies between the lower and upper envelopes
\begin{equation}
    \underline a(x)=a_0(x)-\varepsilon,\qquad
    \overline a(x)=a_0(x)+\varepsilon,\qquad
    S_a=\|a_0\|_\infty+\varepsilon=1.501.
    \label{eq:certified-envelope}
\end{equation}
Because our one-dimensional quadrature bounds depend on how rapidly the input can vary between quadrature nodes, the certified input set for 1D Poisson and 1D Helmholtz also bounds the first and second spatial derivatives:
\begin{equation}
    \mathcal A_P
    =
    \left\{
    a\,\vert \,
    \underline a(x)\le a(x)\le\overline a(x)
    \ \forall x\in[0,1],\quad
    |a'(x)|\le M_a,\quad
    |a''(x)|\le K_a
    \right\},
    \label{eq:experiment-input-set}
\end{equation}
where $M_a = 4.008\pi \approx 12.592$ and $K_a = 12.256\pi^2 \approx 120.962$ combine the derivatives of $a_0(x)$ with the maximum derivatives of the Fourier perturbation family in~\eqref{eq:attack-family}. For 2D Poisson and space--time Burgers, our two-dimensional linearized propagation (Appendix~\ref{app:2d-linearization}) integrates the perturbation directly against the finite Fourier modes of the network, so no derivative bound is required on the perturbation.

\begin{table}[H]
\centering
\footnotesize
\setlength{\tabcolsep}{4pt}
\caption{Input specifications and quadrature grid resolutions used for certification.}
\label{tab:certified-inputs}
\begin{adjustbox}{max width=\linewidth,center}
\begin{tabular}{llccccl}
\toprule
Benchmark & Nominal input $a_0$ & Radius $\varepsilon$ & $|a'|\le M_a$ & $|a''|\le K_a$ & Robustness grid $Q$ & Residual / query grid \\
\midrule
1D Poisson & $\sin(2\pi x)+\tfrac12\cos(4\pi x)$ & $10^{-3}$ & $4.008\pi$ & $12.256\pi^2$ & $2^{20}$ & $2^{25}$ \\
1D Helmholtz & $\sin(2\pi x)+\tfrac12\cos(4\pi x)$ & $10^{-3}$ & $4.008\pi$ & $12.256\pi^2$ & $2^{20}$ & $2^{25}$ \\
2D Poisson & $\sin(\pi x)\sin(\pi y)+\tfrac12\sin(2\pi x)\sin(2\pi y)$ & $10^{-3}$ & --- & --- & $2048\times 2048$ & $16384\times 16384$ \\
Burgers & $\sin(2\pi x)+\tfrac12\cos(4\pi x)$ & $10^{-3}$ & --- & --- & $4096\times 4096$ & $16384\times 4096$ \\
KS rollout & $\sin(x/3)+\tfrac12\cos(2x/3)$ & $0$ & --- & --- & --- & $1\times 64$ panels (adaptive) \\
SFNO (ERA5) & ERA5 spherical harmonics ($\ell\le 239$) & $0$ & --- & --- & --- & $262{,}144\times 196{,}608$ \\
\bottomrule
\end{tabular}
\end{adjustbox}
\end{table}

\subsection{Empirical witness search}
\label{app:attack-details}

To evaluate how tight our certified robustness bounds are, we compare each upper bound against a concrete lower bound obtained by searching for pairs of admissible inputs that produce large output differences (reported as \emph{Empirical width} in Table~\ref{tab:main-results}).

We search over smooth perturbations of the form $a = a_0 + \delta_c$, where
\begin{equation}
    \delta_c=\varepsilon\sum_{b=1}^{B}c_b\phi_b,
    \qquad \sum_{b=1}^{B}|c_b|\le1,
    \label{eq:attack-family}
\end{equation}
and each basis function satisfies $\|\phi_b\|_\infty\le 1$ so that $\|\delta_c\|_\infty\le\varepsilon$ everywhere. In one dimension (1D Poisson, 1D Helmholtz, and Burgers), the basis consists of the constant function together with sine and cosine modes $m=1,\ldots,16$ ($B=33$); for 1D Poisson and Helmholtz, mode $m$ is scaled by $\min(1,4/m)$ so that every candidate satisfies the derivative bounds in~\eqref{eq:experiment-input-set}. In 2D Poisson, the basis consists of all products of the constant and the first four sine and cosine modes in each coordinate ($B=81$).

For each benchmark, we evaluate $4{,}096$ candidate perturbations on the native grid (the $2B$ signed basis functions $\pm\phi_b$, followed by $4{,}096-2B$ random combinations drawn uniformly from $\sum_b |c_b|\le 1$) and identify the two inputs whose outputs differ the most. For the grid-level rows in Table~\ref{tab:main-results}, this sampled difference is reported directly. For the function-space rows, we pass the two selected inputs through our continuum verifier (with $\varepsilon=0$) to obtain rigorous intervals $[l_1, u_1]$ and $[l_2, u_2]$ enclosing their true continuum outputs at the query point. Their separation is then guaranteed to be at least $\max(0,\, l_1 - u_2,\, l_2 - u_1)$, providing a certified lower bound on the true worst-case output variation.
\end{document}